\documentclass[lettersize,journal]{IEEEtran}

\usepackage{amsmath,amsfonts}
\usepackage{algorithmic}
\usepackage{algorithm}
\usepackage{array}
\usepackage[caption=false,font=normalsize,labelfont=sf,textfont=sf]{subfig}
\usepackage{textcomp}
\usepackage{stfloats}
\usepackage{url}
\usepackage{verbatim}
\usepackage{graphicx}
\usepackage{cite}

\usepackage{soul}
\usepackage{multirow}
\usepackage{makecell}
\usepackage{booktabs}
\usepackage{color}

\begin{document}
\title{Cross-Video Contextual Knowledge Exploration and \\ Exploitation for Ambiguity Reduction in \\ Weakly Supervised Temporal Action Localization}
\author{Songchun Zhang, and Chunhui Zhao,~\IEEEmembership{Senior Member,~IEEE}
\thanks{This work is supported by the National Natural Science Foundation of China (No. 62125306) and the Major Science and Technology Infrastructure Project of Zhejiang Laboratory (Large-scale experimental device for information security of new generation industrial control system). \textit{(The corresponding author is Chunhui Zhao).}}
\thanks{Songchun Zhang and Chunhui Zhao are with the College of Control Science and Engineering, Zhejiang University, Hangzhou 310027, China (e-mail: songchunzhang@zju.edu.cn; chhzhao@zju.edu.cn).}
}

\markboth{IEEE TRANSACTIONS ON CIRCUITS AND SYSTEMS FOR VIDEO TECHNOLOGY}
{Shell \MakeLowercase{\textit{et al.}}: A Sample Article Using IEEEtran.cls for IEEE Journals}

\IEEEpubid{\begin{minipage}{\textwidth}\ \\[30pt] \centering
Copyright \copyright 20xx IEEE. Personal use of this material is permitted. \\
However, permission to use this material for any other purposes must be obtained from the IEEE by sending an email to pubs-permissions@ieee.org.
\end{minipage}}

\maketitle
\begin{abstract}
Weakly supervised temporal action localization (WSTAL) aims to localize actions in untrimmed videos using video-level labels.
Despite recent advances, existing approaches mainly follow a \textit{localization-by-classification} pipeline, generally processing each segment individually, thereby exploiting only limited contextual information.
As a result, the model will lack a comprehensive understanding (e.g. appearance and temporal structure) of various action patterns, leading to ambiguity in classification learning and temporal localization.
Our work addresses this from a novel perspective, by exploring and exploiting the cross-video contextual knowledge within the dataset to recover the dataset-level semantic structure of action instances via weak labels only, thereby indirectly improving the holistic understanding of fine-grained action patterns and alleviating the aforementioned ambiguities.
Specifically, an end-to-end framework is proposed, including a \textit{Robust Memory-Guided Contrastive Learning} (RMGCL) module and a \textit{Global
Knowledge Summarization and Aggregation} (GKSA) module.
First, the RMGCL module explores the contrast and consistency of cross-video action features, assisting in learning more structured and compact embedding space, thus reducing ambiguity in classification learning.
Further, the GKSA module is used to efficiently summarize and propagate the cross-video representative action knowledge in a learnable manner to promote holistic action patterns understanding, which in turn allows the generation of high-confidence pseudo-labels for self-learning, thus alleviating ambiguity in temporal localization.
Extensive experiments on THUMOS14, ActivityNet1.3, and FineAction demonstrate that our method outperforms the state-of-the-art methods, and can be easily plugged into other WSTAL methods.

\end{abstract}


\section{Introduction}
\begin{figure}[t!]
    \centering
    \includegraphics[width=0.38\textwidth]{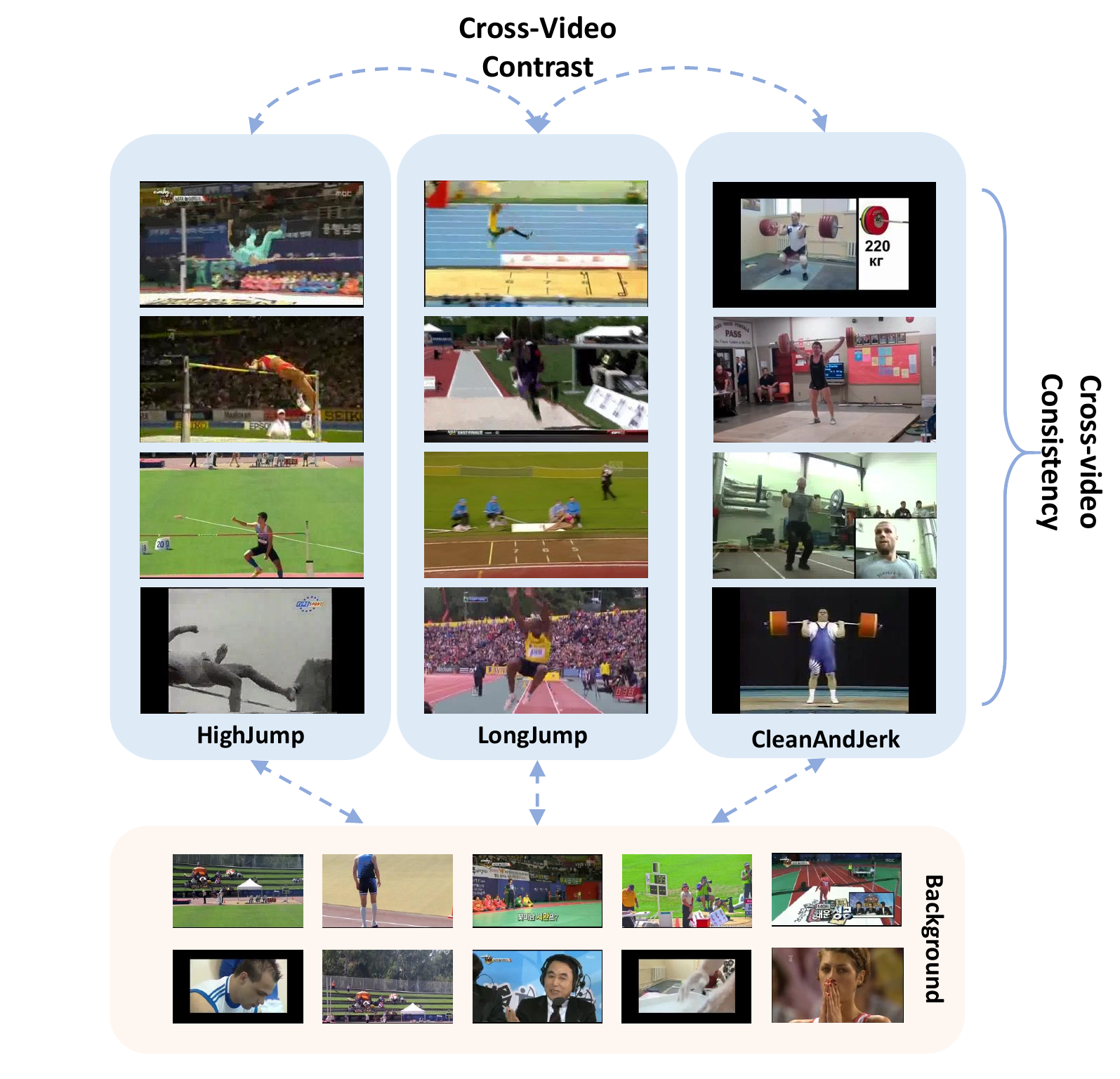}
    \caption{\textbf{An illustration of cross video correlation.} Our approach performs dataset-level action contextual mining to extract rich contextual knowledge from all available training video segments, rather than from individual segments. This enables our model to gain a holistic understanding of action patterns, thus alleviating the ambiguities in the pipeline.}
    \label{fig:first}
    \vspace{-7mm}
\end{figure}

\IEEEPARstart{T}{emporal} action localization aims to localize action instances in the untrimmed video along the temporal dimension, which has a widespread application in different scenarios.
Some existing methods are trained in a fully supervised manner. 
However, as obtaining annotations with accurate action boundaries for extensive video data is an expensive and time-consuming task, such methods are not suitable for real-world applications.
To solve the above problem, the weakly-supervised temporal action localization (WSTAL) method has been proposed.
This method has gained popularity in recent years because it only uses video-level annotations in the training process, which reduces the dependence on expensive annotations.

Inspired by class activation map (CAM)~\cite{zhou2016learning} in object recognition, the approaches~\cite{huang2021foreground,huang2022weakly,leeBackgroundSuppressionNetwork2020,stpn,feng2022bias} based on localization-by-classification pipeline have achieved success on WSTAL task via the adoption of temporal class activation maps (TCAMs).
The whole process involves training an action classification model based on video-level action labels, computing TCAMs based on the classification model, and post-processing the TCAMs for final action proposals.
%
Therefore, it can be observed that the quality of the TCAMs determines the upper bounds of the models.
To improve the quality of obtained TCAMs, many approaches have been proposed, e.g., metric learning-based approaches~\cite{gan2015devnet, huangRelationalPrototypicalNetwork2020, zhangCoLAWeaklySupervisedTemporal2021, li2022dcc, zhou2022object}, attention-based approaches~\cite{leeBackgroundSuppressionNetwork2020,islamHybridAttentionMechanism2021,yangUncertaintyGuidedCollaborative2021,quACMNetActionContext2021,huangForegroundActionConsistencyNetwork2021,huang2021foreground,huang2022weakly, wang2023two, zeng2019graph}, and pseudo-label-based methods~\cite{huang2022weakly,li2022dcc,luo2020emmil,zhaiTwoStreamConsensusNetwork2020,shouAutoLocWeaklySupervisedTemporal2018,heASMLocActionawareSegment2022, ju2023distilling}.

Despite recent significant advances, we observe that there are two pending issues in the localization-by-classification pipeline, i.e. \textbf{ambiguous classification learning} and \textbf{localization ambiguity}.
Specifically, the above issues can be attributed to two main aspects:
(1) For video-level classification, it is generally sufficient to recognize the discriminative segments of an action instance. 
As a result, the classifiers tend to focus more on feature learning of these segments and ignore those segments that are not prominent in this pipeline.
However, ignoring non-prominent segments can directly lead to incomplete action localization, which in turn affects the overall localization performance. 
In addition, the action pattern information embedded in them is lost, leading to low-quality feature representations and ambiguous classification learning.
(2) Under weakly supervised settings, it is challenging for learned classifiers to accurately distinguish actions from backgrounds~\cite{quACMNetActionContext2021,leeBackgroundSuppressionNetwork2020} (i.e., non-action instance frames) due to the lack of frame-level annotations for action instances, which leads to localization ambiguity.

To address the aforementioned ambiguities, we find that leveraging the inherent constraints present in intra- and inter-video contexts can enhance the effectiveness of weakly supervised learning.
Specifically, video-level labels not only provide information about the action categories in individual videos, but also reveal the semantic structure of all actions in the dataset.
For example, there are multiple instances of actions in the dataset that are semantically similar but visually distinct (e.g., \textit{Clean And Jerk} in Fig.~\ref{fig:first}). 
On the other hand, instances of distinct actions (such as \textit{High Jump} and \textit{Long Jump}), are semantically dissimilar despite exhibiting some visual similarities.
By exploiting this prior knowledge, we can gain a more comprehensive understanding of action patterns, thereby reducing ambiguities in the pipeline and improving the performance of current WSTAL methods.


\begin{figure}[t!]
    \centering
    \includegraphics[width=0.5\textwidth]{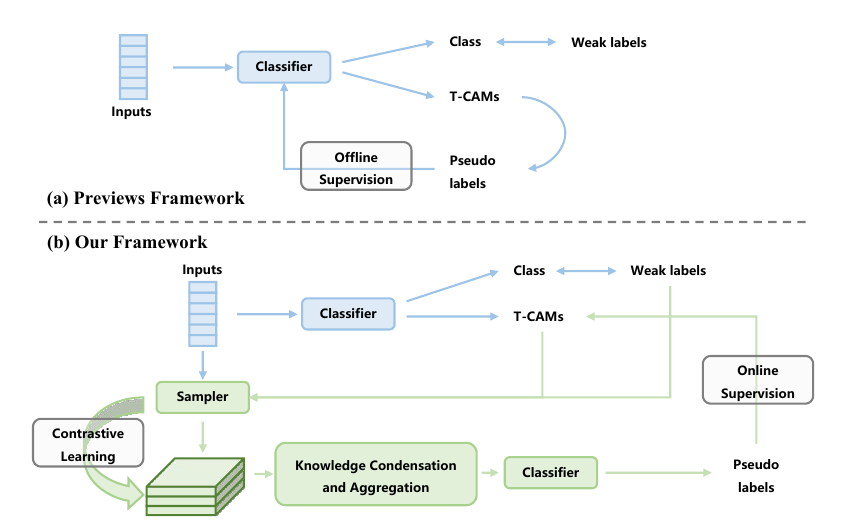}
    \caption{\textbf{The difference between our approach and the previous pseudo-label-based approach.} Previous methods typically use an offline approach to generate pseudo-labels. In our approach, however, we generate high-quality pseudo labels by gathering knowledge of various action patterns to provide online supervised signals for the main branch.}
    \label{fig:brief_pipeline}
    \vspace{-7mm}
\end{figure}


Motivated by the above analysis, we propose a Robust Cross-Video Contrast and Aggregation Learning framework, as shown in Fig.~\ref{fig:brief_pipeline}, to maximize the exploitation of representative action knowledge across the entire video dataset, aiming at holistic action patterns understanding to alleviate ambiguity in classification learning and temporal localization.
%
Specifically, our framework consists of two core strategies, i.e., the \textit{Robust Memory-Guided Contrastive Learning} (RMGCL) and the \textit{Global Knowledge Summarization and Aggregation} (GKSA):
(1) The RMGCL module is designed for compact and reliable feature learning from noisy and weak action instance labels.
Concretely, a continuously updated memory bank is first constructed to store a vast and diverse collection of action patterns (including prominent and non-prominent segments) observed in the training dataset, which subsequently provides support for the exploration of the action semantic structure of the entire dataset, and can further facilitate learning for non-prominent segments.
Then, for each input action embedding, the RMGCL module forces the network to adapt its embedding closer to the memory embeddings that belong to the same action category, while pushing away those embeddings that belong to a different action category.
In particular, to address the issue of noise contrastive learning under weak label settings, we design a symmetric noise-tolerant contrastive loss, which theoretically guarantees the noise robustness of contrastive learning.
In summary, the RMGCL module can help the model learn a more structured and compact embedding space, thus alleviating the ambiguity in classification learning.
(2) The GKSA module is designed for efficiently gathering global contextual knowledge to provide a comprehensive understanding of action patterns and yield high-quality segment-level pseudo labels for self-teaching.
Specifically, we further summarize and condense the features in the memory bank in a learnable manner using a set of latent codewords and a transformer-based knowledge summarization module.
To alleviate temporal localization ambiguity as well as improve localization precision for non-prominent segments, the cross-video contextual knowledge is propagated into current embeddings via a non-parametric cross-attention module to improve the holistic understanding of action patterns, which in turn generates pseudo-labels to provide high-quality segment-level supervision for the main classification branch.


\par


Our contributions are as follows:
\begin{itemize}
    \item In this research, we analyze two key issues in the existing WSTAL models, namely the ambiguous classification learning and localization ambiguity, and propose an end-to-end framework to address both issues from the perspective of cross-video action context mining.
    \item We propose two core modules, namely RMGCL and GKSA, to jointly address the above issues.
    Specifically, the RMGCL modules equipped with a continuously updated memory bank can alleviate ambiguity in classification learning by learning more structured and compact embedding space from noisy action features.
    The GKSA modules can efficiently gather cross-video representative knowledge to generate high-quality frame-level pseudo-labels for self-teaching, which in turn helps to mitigate localization ambiguity.   
    \item Extensive experiments have been conducted on three widespread benchmarks to demonstrate the effectiveness of our method.
    Compared with previous state-of-the-art methods, our method improves average mAP by $1.8\%$, $1.4\%$, and $1.0\%$ on THUMOS14~\cite{idreesTHUMOSChallengeAction2017}, ActivityNet1.3~\cite{caba2015activitynet}, and FineAction~\cite{liu2022fineaction} datasets, respectively. And it can be easily applied to other methods to improve their performance.
\end{itemize}

\section{Related Work}
In this section, we review previous work that provides the motivation for our proposed approach.
These works can be divided into two groups: weakly-supervised temporal action localization, and contrastive representation learning.
We then review each of the previous works separately.


\subsection{Weakly-Supervised Temporal action localization}
Fully supervised temporal action localization~\cite{lin2018bsn,ssn} typically leverages accurate temporal annotations for model training, which is difficult to obtain in practical applications. 
To overcome this limitation, the WSTAL task has drawn significant attention in
recent years by exploiting only video-level supervision for model training.
\par
Many methods have been proposed in recent years, which mainly employ a localization-by-classification pipeline.
Specifically, these methods~\cite{wang2017untrimmednets,a2cl-pt,narayan20193c,paulWTALCWeaklySupervisedTemporal2018,leeBackgroundSuppressionNetwork2020,sunSlowMotionMatters2022,wang2021exploring,islam2021hybrid, ju2022adaptive, zeng2019breaking, yang2021multi} first optimize a video-level classifier using weak labels and then generate frame-level TCAMs, after which a threshold can be set to localize the action instances.
BaS-Net~\cite{leeBackgroundSuppressionNetwork2020} introduces an auxiliary frame-level attention branch to suppress the activation of the background.
ACM-Net~\cite{quACMNetActionContext2021} models action instances, action backgrounds, and action contexts simultaneously by constructing three parallel frame-level attention branches.
Sun \textit{et al}~\cite{song2022slow} focuses on improving temporal action localization results by mining and utilizing slow-motion features in video segments as complementary information.
MSA-Net~\cite{yang2021multi} proposes a framework to explore both the global structure information of a video and the local structure information of actions for robust localization.
%
However, the lack of accurate frame-level annotations can lead to problems of localization ambiguity and ambiguous classification learning.
To alleviate this, erasing-based methods~\cite{zeng2019breaking, kumar2017hide, min2020adversarial} carefully design adversarial erase strategies, which find less discriminative regions by erasing the most discriminant region.
The pseudo-label-based approaches~\cite{pardoRefineLocIterativeRefinement2021, luo2020emmil, huang2022weakly, li2022dcc, heASMLocActionawareSegment2022} attempt to use pseudo-TCAMs to provide segment-level supervision to alleviate the ambiguity.
RefineLoc~\cite{pardoRefineLocIterativeRefinement2021} is the first method to incorporate pseudo-labels into the WSTAL task.
EM-MIL~\cite{luo2020emmil} and RSKP~\cite{huang2022weakly} introduce the generation of pseudo-labels into an expectation maximization framework.
UGCT~\cite{yangUncertaintyGuidedCollaborative2021} proposes an uncertainty-guided training strategy with uncertainty estimation for the generation of pseudo-labels.
AMS~\cite{ju2022adaptive} constructs a two-branch framework and proposes a mutual location supervision, which forces each branch to use pseudo-labels obtained from the other branch.
However, most of them process each segment individually, using only limited contextual information.
This results in the model lacking a holistic understanding of action patterns, which makes it challenging to effectively resolve the issues of ambiguous classification learning and localization ambiguity.
Our work is mostly related to~\cite{wang2021exploring}, which proposes to operate at the sub-action granularity to explore cross-video relationships. 
The learning of sub-action representation is guided by two-branch consistency loss.
Different from them, we summarize knowledge from a massive global memory bank via a set of latent codewords.
Thus, we can extract reliable cross-video contextual to give more instructive guidance to the entire learning process.

%
\begin{figure*}[t!]
    \centering
    \includegraphics[width=0.95\textwidth]{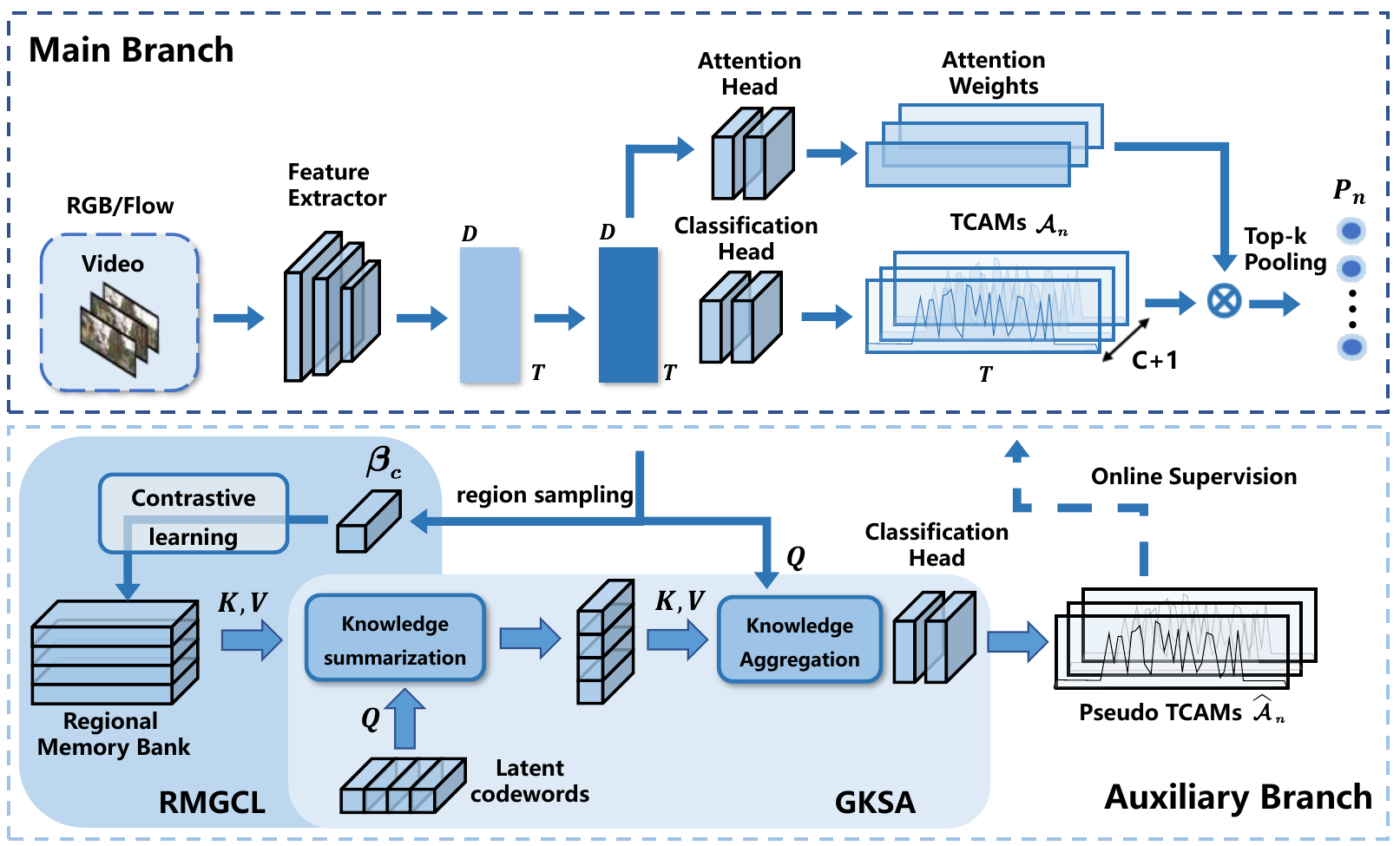}
    \caption{\textbf{Pipeline of our proposed architecture.}  The model consists of two main parts, i.e., the main branch and the auxiliary branch. For the main branch, we adopt a similar design as~\cite{quACMNetActionContext2021} as our baseline model to generate TCAMs and video-level probability distribution $\boldsymbol{p}_n$. In the auxiliary branch, we first construct a dataset-level memory for storing massive representative segment features $\beta_c$ and obtain a more compact and reliable feature representation via the RMGCL module. We then design the GKSA module to gather cross-video contextual knowledge to promote holistic action patterns understanding, which in turn allows the generation of high-confidence pseudo-labels for self-teaching.}
    \label{fig:overall_pipeline}
    \vspace{-4mm}
\end{figure*}

\subsection{Contrastive Representation Learning}
Contrastive representation learning~\cite{chen2021exploring,caron2018deep,chen2020simple,he2020momentum,wu2018unsupervised,liu2022multi} is an unsupervised data representation learning method that trains a model to distinguish between similar and dissimilar examples.
The aim is to learn a feature space in which similar examples are next to each other and dissimilar examples are far apart.
In addition, supervised contrast learning~\cite{khosla2020supervised} is proposed for image recognition and semantic segmentation. The method extends the self-supervised setup by comparing all sample sets from the same category as positives with negatives from other categories.
Recently, contrastive learning has also been used in video representation learning~\cite{liu2022semi,tao2022improved, chen2022consistent,huang2021self}, action recognition~\cite{xu2022x, shu2022multi, xu2023pyramid, xu2023spatiotemporal}.
Chen \textit{et al}~\cite{chen2022consistent} introduce intra-video contrastive learning that further distinguishes intra-video actions to alleviate temporal collapse. However, we utilize both inter- and intra-video contextual knowledge to reduce ambiguity in classification learning.
SDS-CL~\cite{xu2023spatiotemporal} proposes three contrasting loss functions to enhance the ability of the model to learn joint and motion features.
MAC-Learning~\cite{shu2022multi} proposes to learn multi-granularity representations between three granularities including local view, contextual view, and global view.
X-CAR~\cite{xu2022x} proposes a contrastive learning framework to obtain rotate-shear-scale invariant features by
learning augmentations and representations of skeleton sequences.
PSP-Learning~\cite{xu2023pyramid} proposes to jointly learn body-level, part-level, and joint-level representations of joint and motion data via contrastive learning.
In addition, some WSTAL methods~\cite{zhangCoLAWeaklySupervisedTemporal2021,luoActionUnitMemory2021,li2022dcc} also employ contrastive learning to improve the discriminability of action feature.
CoLA~\cite{zhangCoLAWeaklySupervisedTemporal2021} argues that learning by comparing can help identify ambiguous segments in videos.
AUMN~\cite{luoActionUnitMemory2021} uses a memory bank to store the appearance and motion information of action units and their corresponding classifiers to achieve more complete localization results.
DCC~\cite{li2022dcc} design a denoised contrastive algorithm to enhance the feature discrimination and alleviate the noisy issue of segment-level labels.

However, existing methods often have difficulty in obtaining the desired structured and compact embedding space through original segment-level contrast learning.
This is mainly due to the ambiguity of action labels in weakly supervised settings, leading to numerous invalid positive and negative pairs.
These conflicting objectives can hinder the convergence of contrastive learning-based models.

\section{Method}

In this section, we first illustrate the problem definition of the WSTAL task. 
Then we will introduce our baseline setup (i.e., the structure of our main branch).
This is followed by introducing the \textit{Memory-Guided Contrastive Learning} and \textit{Global Knowledge Summarizat} strategies.
Finally, we give the overall training objectives and the relevant details of the inference process.
The overall structure of our model is shown in Fig.~\ref{fig:overall_pipeline}.

\subsection{Problem Definition}
\label{sec:problem_def}
In WSTAL task, we have a bag of $N$ training videos $\{V_n\}^N_{n=1}$ being annotated with their action categories $\{\mathbf{y}_n\}^N_{i=n}$, where $\mathbf{y}_n$ is a binary vector indicating the presence or absence of a certain class of actions in video $V_n$.
Noted that multiple categories of actions can exist simultaneously in the same video. 
Each video $V$ can be divided into a series of non-overlapping segments: $V = \{v_t\}^{T}_{t = 1}$, where $T$ is the number of segments in the video, and $v_t$ denotes the t-th segment in the video $V$.
Our main purpose is to figure out a set of action instances $\{(c,q,t_s,t_e)\}$ in the test video by leveraging the weakly labeled training dataset $\{V_n,\mathbf{y}_n\}^N_{n=1}$, where $c$ denotes the predicted class, $q$ indicates the confidence score, $t_s$ and $t_e$ indicate the start time and end time of current action instance, respectively.

\subsection{Baseline Setup}
\label{sec:main_branch}

In this section, we will briefly introduce the baseline of our method, which contains two components, i.e. \textit{Feature Extraction and Embedding} module and \textit{Classification Head} module.

\subsubsection{\textbf{Feature Extraction and Embedding}}
Given a training video, we first divide it into a series of non-overlapping segments. 
Then, we use a pre-trained backbone network Inflated 3D (I3D)~\cite{I3D} to extract the RGB and motion feature $\boldsymbol{X}=\left\{\boldsymbol{x}_{\boldsymbol{i}}\right\}_{i=1}^T$, where $\boldsymbol{x}_i \in \mathbb{R}^D$.
Since the I3D backbone is pre-trained on the Kinetics-400 for action recognition, we then encode them into latent embeddings $\mathbf{F}=\left\{\boldsymbol{f}_i\right\}_{i=1}^T \in \mathbb{R}^{T \times D}$ through temporal convolution, where $T$ is the number of segments of a video.
The process is described as follows:

\begin{equation}
    \mathbf{F} = \sigma (conv(\boldsymbol{X}; \theta_{embed})),
\end{equation}
where $\sigma$ indicates the ReLU function, the $conv(\cdot)$ denotes temporal convolution layers and $\theta_{embed}$ indicates the trainable parameters of convolution layers.

\begin{figure}[t!]
    \centering
    \includegraphics[width=0.5\textwidth]{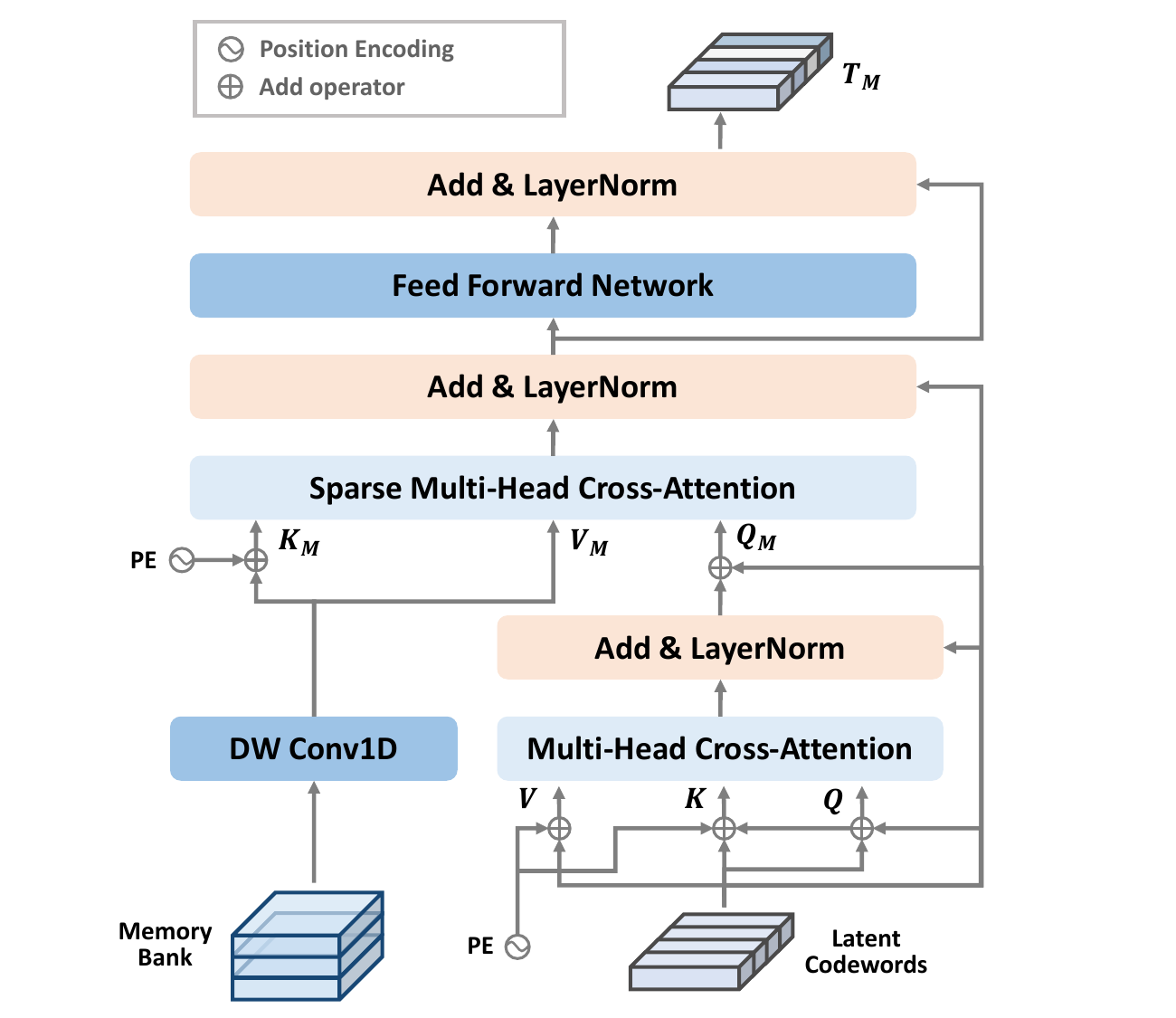}
    \caption{\textbf{The architecture of Knowledge Summarization module.} We adopt a DETR-like~\cite{detr} architecture. By setting up a set of randomly initialized latent codewords, representative action knowledge can be efficiently summarized from the memory bank in a learnable manner. At the same time, this process facilitates the information interaction between the different action instances, which further enhances holistic understanding.}
    \label{fig:summarazation}
    \vspace{-4mm}
\end{figure}

\subsubsection{\textbf{Classification Head}}
The Classification Head is typically employed to project segment features to TCAMs, which are then used for subsequent video-level classification and segment-level action localization tasks.
The process of obtaining TCAMs can be described by the following equation:

\begin{equation}
\label{eq:cam}
    \mathcal{A} = \mathcal{F}_{CAM}(\mathbf{F}; \theta_{cam}),
\end{equation}

\noindent where $\mathcal{F}_{CAM}$ and $\theta_{cam}$ are the projection layer and learnable parameters of the projection layer, respectively. The projection layer is implemented as a single fully connected layer.


Then, a three-branch temporal attention module~\cite{quACMNetActionContext2021} is adopted to explicitly model action, action-context, and background $\mathcal{W}=\{(\mathcal{W}^{ins}(i),\mathcal{W}^{con}(i),\mathcal{W}^{back}(i))\}^T_{i=1} \in \mathbb{R}^{T \times 3}$, respectively.
Specifically, the three-branch segment-level attention can be modeled by the following equation:
\begin{equation}
    \mathcal{W} = \operatorname{Softmax}(\mathcal{F}_{att}(\mathbf{F};\theta_{att})),
\end{equation}
where $\mathcal{F}_{att}$ and $\theta_{att}$ are the attention head module and the learnable parameters of the attention head. 
The attention head module is implemented as a single temporal convolutional layer with a kernel size of 3.

After generating the attention weights, the attention-weighted TCAMs can be obtained as follows:

\begin{equation}
    \mathcal{A}^m = \mathcal{W}^m \otimes \mathcal{A}, m \in \{ins, con, back\}
\end{equation}

Following Multi-Instance Learning~\cite{MIL}, the segment-level classification scores can be aggregated to generate a video-level class score $p_{c}^m, m\in\{ins, con, back\}$ for each segment type by applying top-k pooling and softmax on attention-weighted TCAMs.
By leveraging segment-wise cross-entropy loss~\cite{heASMLocActionawareSegment2022, leeBackgroundSuppressionNetwork2020, huang2021foreground, huang2022weakly}, the objective function can be derived as follows:

\begin{equation}
    \mathcal{L}_{c l s}^{m}=\sum_{c=1}^{C+1}-y_{c}^{m} \log \left(p_{c}^{m}\right), m \in \{ins, con, back\}
\end{equation}
where $p_{c}$ indicates video-level action probability distribution, $y_{c}$ refers to the ground-truth video action probability distribution. Specifically, $y^{ins} = [y_c = 1, y_{C+1} = 0]$, $y^{con} = [y_c = 1, y_{C+1} = 1]$ and $y^{back} = [y_c = 0, y_{C+1} = 1]$ for action instance, action context, and background, respectively.

\subsection{Robust Memory-Guided Contrastive Learning}
\label{sec:Regional}
Recent work~\cite{huangForegroundActionConsistencyNetwork2021, quACMNetActionContext2021,leeBackgroundSuppressionNetwork2020} mainly use temporal attention on untrimmed videos for action classification and localization.
However, such approaches do not account for the contrast and consistency of action features across videos.

To fully exploit the inductive bias information provided by the entire dataset, we first construct a non-parametric regional feature memory for storing and updating representative pseudo action region features. 
Subsequently, a robust contrast learning strategy is employed to mine the consistency and contrast of cross-view actions from noisy features.

\par

\subsubsection{\textbf{Memory Bank Construction}}

For each training video sample, the regional representative feature segments in the current video can be obtained using the predicted TCAMs and video-level weak labels.
Specifically, the regional representative feature segments of a specific action class $c$ in the current video can be derived by the following equation:

\begin{equation}
    \boldsymbol{\beta}_c = \frac{\sum_{t=1}^T \boldsymbol{\mu}_c (t) \otimes \boldsymbol{f}_t}{\sum_{t=1}^T \boldsymbol{\mu}_c (t)},
\end{equation}

\noindent where $\boldsymbol{\mu}_c (t)$ indicates whether segment $t$ is involved in the calculation of regional representative features of category $c$.
This can be obtained according to the following equation:

\begin{equation}
    \boldsymbol{\mu}_c = \left[\mathcal{A}^{ins}[:,c] > \zeta\right] \in \mathbb{R}^T,
\end{equation}

\noindent where {$\zeta$} is manually set to 0.75 for filtering low-confidence segments, and $\mathcal{A}^{ins}$ indicates the attention-weighted TCAMs for action instance segments we obtained in the previous step. 

After obtaining the class-wise representative segment feature $\boldsymbol{\beta}_c$, we then construct a non-parametric regional memory bank $\mathbf{M} \in \mathbb{R}^{(C + 1) \times q \times D}$ to store and update the dataset-level feature dynamically, where $q$ indicates the feature queue length of each action class.
Note that the features of the background action classes are also stored, as we want to encourage the separation between classes, as well as between background and foreground, which will be discussed in Section~\ref{sec:ablation_study}.
Following the strategy mentioned in~\cite{he2020momentum}, the current feature $\boldsymbol{\beta_c}$ of action class $c$ will be progressively updated to the memory bank as follows:

\begin{equation}
    \mathbf{m}_c \leftarrow (1 - \alpha) \mathbf{m}_c + \alpha \boldsymbol{\beta}_c,
\end{equation}
where $\alpha$ refers to the momentum coefficient, and $\mathbf{m}_c$ denotes the feature stored in the memory queue of the action class $c$.

In addition, to improve the overall quality of the representative features, we are required to introduce video-level weak labels to filter the incoming features in the updating process.

\par



\subsubsection{\textbf{Memory-Guided Contrastive Learning}}
To encourage cross-video feature contrast and consistency, we introduce a robust feature contrastive learning strategy. 
Our major goal is to enforce the contrast between features of different action classes while encouraging the similarity within features of the same action class.

The positive/negative sample pairs are constructed from the action segments within and across the videos.
Specifically, given a representative feature $\boldsymbol{\beta}_c$, the positive sample set $\mathcal{P}_c$ consists of features in the memory that are in the same class as the given feature.
On the other hand, the negative sample set $\mathcal{N}_c$ consists of two parts, i.e., features of a different class from the given feature and background features.
However, due to the weak and noisy labels, it is challenging to obtain robust representations through the contrastive learning strategy in our task.
To alleviate this problem, inspired by~\cite{zhang2018generalized}, we adopt a noise-tolerant loss to cope with noisy and weak action labels.


\begin{equation}
\begin{aligned}
& \mathcal{L}_{\text {cont }}\left(\beta_c\right)=-\sum_{r_c^{+} \in \mathcal{P}_c} \frac{1}{q} e^{q \cdot\left\langle\beta_c, r_c^{+}\right\rangle / {\tau}} \\
& +\sum_{r_c^{+} \in \mathcal{P}_c} \frac{1}{q}\left[\lambda \cdot\left(e^{\left\langle\beta_c, r_c^{+}\right\rangle / {\tau}}+\sum_{{r_c^{-}} \in \mathcal{N}_c} e^{\left\langle\beta_c, r_c^{-}\right\rangle / {\tau}}\right)\right]^q
\end{aligned}
\end{equation}

\noindent where $\lambda \in(0,1]$ is the density weighting parameter that controls the ratio between positive and negative pairs, {$q$} is a key parameter that can be used to balance the convergence and robustness of the algorithm, $\tau$ is a temperature hyper-parameter used to adjust the distribution of distance, and $\left \langle \cdot, \cdot \right \rangle$ indicates the cosine similarity metric.

We then analyze the effect of the parameter {$q$}.
When $q$ is close to $0$, the algorithm performs hard positive mining, providing faster convergence in the absence of noise.
However, in the presence of noise, the algorithm will incorrectly place higher weights on the false positive pairs because noisy samples tend to cause larger losses, which may prevent the algorithm from converging.
In contrast, when $q$ is close to $1$, the algorithm can achieve robustness, since the loss function satisfying the symmetric form has a strong theoretical guarantee against noise labels, as described in~\cite{ghosh2015making}.

\subsection{Global Knowledge Summarization and Aggregation}
\label{sec:knowledge}

\subsubsection{\textbf{Global Knowledge Summarization}}
In this section, we first attempt to summarize and compress the representative action knowledge stored in the memory bank to obtain a more compact feature representation while smoothing out the noisy information in the memory bank.
Specifically, we construct a set of latent codewords $\mathcal{R} \in \mathbb{R}^{N \times D}$ to summarize the action knowledge stored in memory.
The latent codewords are randomly initialized and shared across the entire training dataset which is updated by back-propagation during the training process.

To adequately fuse and summarize the representative features stored in memory, we design a DETR-like~\cite{detr} architecture to query and aggregate representative action knowledge via a set of learnable latent codewords, as shown in Fig.~\ref{fig:summarazation}.
Specifically, we first use a self-attention module to allow information interaction between the latent codewords and generate the query embedding $\mathbf{Q}_M$ for the regional memory bank, which can be derived as follows:

\begin{equation}
\begin{gathered}
\mathbf{K}=\operatorname{MLP}(\operatorname{Concat}(\mathcal{R}, P E)) \\
\mathbf{V}=\operatorname{MLP}(\operatorname{Concat}(\mathcal{R}, P E)) \\
\mathbf{Q}_M=\operatorname{Self-Attention}(\mathcal{R}, \mathbf{K}, \mathbf{V})
\end{gathered}
\end{equation}

\noindent where $\operatorname{MLP}(\cdot)$ indicates Multi-Layer Perception, and $PE \in \mathbb{R}^{N \times D}$ denotes the positional embeddings.

After obtaining the memory query vector $\mathbf{Q}_M$, we summarize and aggregate the representative action knowledge in the memory efficiently via a sparse cross-attention module.
This process can be illustrated by the following equation.

\begin{equation}
\begin{gathered}
\mathbf{K}_M=\operatorname{DWConv}_{1 \times 1}(\mathbf{M}) \\
\mathbf{V}_M=\operatorname{DWConv}_{1 \times 1}(\mathbf{M}) \\
\mathbf{T}_M=\operatorname{FFN}(\operatorname{Sp-Att}\left(\mathbf{Q}_M, \operatorname{Concat}\left(\mathbf{K}_{\mathrm{M}}, P E\right), \mathbf{V}_{\mathrm{M}}\right))
\end{gathered}
\end{equation}

\noindent where $\mathbf{M}$ refers to the regional memory bank, $\operatorname{DWConv}_{1 \times 1}(\cdot)$ indicates depthwise separable convolution for computational efficiency~\cite{chollet2017xception}, $\operatorname{FFN}(\cdot)$ denotes the feed forward network, and $\operatorname{Sp-Att}(\cdot)$ indicates the sparse cross-attention module.
Noted that the attention and FFN layers use the designs of residual connection and layer normalization of transformers~\cite{transformer}.

Finally, representative feature segments stored in the memory bank can be summarized as $N$ latent codewords.
In general, $N \ll C \times q$. Thus, the potential summaries can provide a more compact representation than the feature memory bank.

\subsubsection{\textbf{Global Knowledge Aggregation}}
After obtaining a summarization of the regional memory bank, we consider designing an efficient method to propagate the knowledge at the dataset level into current video segment features $\mathbf{F}$.
To achieve the above goal, we need to decode and fuse the knowledge compressed in the latent codewords $\mathbf{T}_M$.
Specifically, we view the current segment features as a set of query vectors and aggregate the latent codewords via a non-parametric attention module.
This process can be illustrated by the following equation:

\begin{equation}
    \mathbf{F}^\prime = \operatorname{Softmax} (\mathbf{F} \cdot \mathbf{T}_M^\top / \sqrt{D} ) \cdot \mathbf{T}_M
\end{equation}

\noindent where $\mathbf{F}^\prime$ indicates the enriched feature representation of $\mathbf{F}$.
Note that we do not use parametric attention modules here, as we will discuss in Section~\ref{sec:ablation_study}.
Then we concatenate and fuse $\mathbf{F}^\prime$ and the original feature $\mathbf{F}$ as follow:

\begin{equation}
    \mathbf{F}_{fuse} = \operatorname{FFN}(\operatorname{Concat}(\mathbf{F}^\prime , \mathbf{F}))
\end{equation}

Here, $\mathbf{F}_{fuse}$ not only encodes local information within the current video but also captures global action information across the entire video dataset, thus providing a complete and holistic feature representation for subsequent tasks.
After obtaining holistic segment features, we can use the global contextual information to generate higher quality pseudo-labels with shared parameter classification heads.
Specifically, the pseudo labels can be acquired as follows:
\begin{equation}
    \hat{\mathcal{A}} = \mathcal{F}_{CAM}(\mathbf{F}_{fuse}; \theta_{cam}),
\end{equation}

\noindent where $\hat{\mathcal{A}}$ denotes the generated pseudo labels.
The high-quality pseudo TCAMs can further provide segment-level supervision for origin TCAMs $\mathcal{A}$, which can improve the completeness of temporal action suggestions and alleviate the conflict between classifiers and detectors.

\subsubsection{\textbf{Discussion}}
\textit{Why can $\hat{\mathcal{A}}$ generated by GKSA be used as a pseudo-label? What about the confidence of $\hat{\mathcal{A}}$?}
In our framework, a large-scale representative memory bank is constructed, which contributes significantly to the localization task but can only be used during training, and thus can also be viewed as \textbf{privileged features}~\cite{xu2020privileged, vapnik2015privilegedlearning}.
To make full use of the inter- and intra-video information (privileged features) stored in the memory bank and to improve the quality of the pseudo-labels, we adopt the privileged knowledge distillation framework~\cite{xu2020privileged} for pseudo-label learning. 
Specifically, the teacher network (classification head in auxiliary branch) and the student network (classification head in main branch) have the same structure, but the input network features are different. 
The teacher network is additionally input with privileged features. 
The knowledge extracted from the teacher network (i.e., the pseudo-labels in our method) is then used to supervise the training of the student network. Since the teacher model contains representative knowledge from multiple sources, it can provide more reasonable guidance to the student model.
We can observe from our quantitative results that the pseudo-label approximates the frame-level action ground truth quite well. 
In addition, in the ablation study, we can see that the mAP metrics are significantly improved after adding pseudo-label supervision. 
In conclusion, we can infer that by combining global contextual information from the memory bank, the pseudo-labels we obtain have high confidence and can indeed provide useful guidance for the main branch.

\subsection{Training Objectives and Inference}
\label{sec:Training}
\subsubsection{\textbf{Pseudo Labels Supervision Loss}}
Although we have previously obtained high-quality pseudo-labels, there are still some inaccurate pseudo-labels.
To deal with the effect of label noise, we guide the model to learn from the noisy pseudo-labels by introducing uncertainty~\cite{kendall2017uncertainties}.
Specifically, referring to the method in~\cite{gawlikowski2023survey, smith2018understanding}, we model the uncertainty of the pseudo-labels by predictive variance. 
Intuitively, after the classifier parameters are fixed, when the outputs of two classifiers are close to each other, the uncertainty is lower, indicating less noise, whereas a larger difference indicates a higher uncertainty and the possible presence of noise in the segments, thus reducing the impact on model training.
In this case, the prediction variance is derived from the Kullback-Leibler divergence of the outputs of the two classifiers, as shown in the following equation:

\begin{equation}
D_t=\sum_{c=1}^{C+1}\left[\mathcal{A}(t,c) \log \left(\frac{\mathcal{A}(t,c)}{\hat{\mathcal{A}}(t,c)}\right)\right]
\end{equation}

\noindent where $D_t$ indicates the uncertainty score at time step $t$, $\mathcal{A}(t) \in \mathbb{R}^{(C+1)}$ indicates the CAM of segment $t$, and $\hat{\mathcal{A}}(t) \in \mathbb{R}^{(C+1)}$
indicates the pseudo-CAM of segment $t$.

Then, combining the uncertainty score with the cross-entropy loss, i.e., limiting the contribution of segments with high uncertainty scores to the loss, the uncertainty-weighted pseudo labels supervision loss can be derived as follows:

\begin{equation}
\mathcal{L}_{p s}=\frac{1}{T}\left[e^{-D_t} \mathcal{L}_{c e}\left(\mathcal{A}(t), \hat{\mathcal{A}}(t)\right)+\rho D_t\right]
\end{equation}

\noindent where $\mathcal{L}_{ce}$ indicates the cross-entropy loss, and $\rho$ denotes the weight of the variance regularization term, which prevents the model from continuously predicting large variances.

\subsubsection{\textbf{Overall Training Objectives}}
Given the current training video feature $\mathbf{F}$ as input.
The main branch and region memory branch are trained with video-level weak labels simultaneously.
The total training objectives can be divided into three parts, i.e., the action classification loss $\mathcal{L}_{cls}$, the contrastive learning loss $\mathcal{L}_{cont}$ and the pseudo supervision loss $\mathcal{L}_{ps}$, which can be derived as follow:

\begin{equation}
\begin{gathered}
    \mathcal{L}_{cls} = \mathcal{L}_{cls}^{ins} + \mathcal{L}_{cls}^{back} + \mathcal{L}_{cls}^{con} \\
    \mathcal{L}_{total} = \mathcal{L}_{cls} + \gamma\mathcal{L}_{ps} + \mu\mathcal{L}_{cont}
\end{gathered}
\end{equation}
where $\gamma$ and $\mu$ are hyper-parameters to balance each part of training objectives.

\subsubsection{\textbf{Inference}}
Inputting a test video, we first predict the TCAMs at the segment level and recognize the action classes present in the video by the video-level action class score distribution.
Note that in this process we can utilize the latent representation $\mathbf{T}_M$ stored in advance.
Then we apply threshold strategy on attention-weighted instance class activate map as prior work~\cite{heASMLocActionawareSegment2022, huang2021foreground} to generate a set of candidate action proposals.
After that, the duplicate proposals are removed by performing non-maximal-suppression (NMS), and the final action localization results are obtained.

\subsubsection{\textbf{Discussion}}
\textit{What about the computational efficiency and storage efficiency of the memory bank?}
As we observe that the representations contained in large-scale memory banks are redundant, and that directly summarizing large-scale representations greatly slows down the learning and inference process. 
Therefore, we first summarize and compress the memory banks via the GSKA module to obtain the global latent representation $T_M$.
During the practical application (inference phase), we delete the memory bank and keep only the compressed global representation $T_M$ to propagate the action knowledge. 
This greatly reduces the cost of maintaining large memory banks during model deployment.
More experimental results about the complexity can be found in Section~\ref{sec:Complexity}.

\section{EXPERIMENTAL RESULTS}

\subsection{Dataset and Evaluation}
In this section, we first validate the effectiveness of our approach on two widely used large benchmark datasets, namely THUMOS14 and ActivityNet.
Then, we further evaluate our method on the FineAction dataset~\cite{liu2022fineaction}, which consists of fine-grained action classes and is thus very challenging for weakly supervised action localization methods.

\subsubsection{\textbf{THUMOS14 Dataset}}
THUMOS14~\cite{idreesTHUMOSChallengeAction2017} is a dataset that has been widely used to evaluate the performance of the WTAL methods.
In this work, we train the model using a subset of THUMOS14 that has frame-level annotations for 20 action categories.
Specifically, we train the model on 200 untrimmed videos from its validation set and evaluate it on 212 untrimmed videos from the test set.
\par

\begin{table*}[!t]
\begin{center}
\caption{Comparisons of temporal localization on THUMOS14 Dataset. UNT and I3D represent UntrimmedNet features and I3D features, respectively. $\dag$ means that the method utilizes additional weak supervisions, action frequency.}
\label{table:THUMOS14}
\scalebox{0.9}{
\begin{tabular}{c|l|c|ccccccc|ccc}
\hline
\hline
\multirow{2}{*}{Supervision} & \multirow{2}{*}{Method} & \multirow{2}{*}{Feature} & \multicolumn{7}{c|}{mAP @ IoU (\%)} & \multirow{2}{*}{\makecell{AVG\\(0.1:0.5)}} & \multirow{2}{*}{\makecell{AVG\\(0.3:0.7)}} & \multirow{2}{*}{\makecell{AVG\\(0.1:0.7)}} \\
\cline{4-10}
&  &  & 0.1 & 0.2 & 0.3 & 0.4 & 0.5 & 0.6 & 0.7 &  &  & \\
\hline
\hline
\multirow{5}{*}{Full}
& SSN~\cite{ssn} (ICCV'17) & - & 60.3 & 56.2 & 50.6 & 40.8 & 29.1 & - & - & 49.6 & - & -\\
& {BSN~\cite{lin2018bsn} (ECCV'18)} & - & - & - & 53.5 & 45.0 & 36.9 & 28.4 & 20.0 & - & 36.8 & - \\
& {BMN~\cite{lin2019bmn} (ICCV'18)} & - & - & - & 56.0 & 47.4 & 38.8 & 29.7 & 20.5 & - & 38.5 & - \\
& {G-TAD~\cite{xu2020gtad} (ECCV'20)} & - & - & - & 53.5 & 45.0 & 36.9 & 28.4 & 20.0 & - & 36.8 & - \\
& {TadTR~\cite{TadTR} (TIP'22)} & I3D & 69.5 & 67.8 & 62.4 & 57.4 & 49.2 & 37.8 & 26.3 & 46.6 & - & - \\
\hline
\multirow{2}{*}{Weak $\dag$}
& STAR~\cite{STAR} (AAAI'19) & I3D & 68.8 & 60.0 & 48.7 & 34.7 & 23.0 & - & - & 47.0 & - & -\\
& 3C-Net~\cite{narayan20193c} (ICCV'19) & I3D & 59.1 & 53.5 & 44.2 & 34.1 & 26.6 & - & 8.1 & 43.5 & - & -\\
\hline
\multirow{20}{*}{Weak}
& UntrimmedNet \cite{wang2017untrimmednets}, (CVPR'17) & - & 44.4 & 37.7 & 28.2 & 21.1 & 13.7 & - & - & 29.0 & - & - \\
& AutoLoc \cite{shou2018autoloc}, (ECCV'18) & UNT & - & - & 35.8 & 29.0 & 21.2 & 13.4 & 5.8 & - & 21.0 & - \\
& STPN~\cite{stpn} (CVPR'18) & I3D & 52.0 & 44.7 & 35.5 & 25.8 & 16.9 & 9.9 & 4.3 & 35.0 & 18.5 & 27.0 \\
& W-TALC~\cite{nguyenWeaklySupervisedAction2018} (ECCV'18) & I3D & 55.2 & 49.6 & 40.1 & 31.1 & 22.8 & - & 7.6 & 39.8 & - & - \\
& BaS-Net~\cite{leeBackgroundSuppressionNetwork2020} (AAAI'20) & I3D & 58.2 & 52.3 & 44.6 & 36.0 & 27.0 & 18.6 & 10.4 & 43.6 & 27.3 & 35.3 \\
& RPN~\cite{huangRelationalPrototypicalNetwork2020} (AAAI'20) & I3D & 62.3 & 57.0 & 48.2 & 37.2 & 27.9 & 16.7 & 8.1 & 46.5 & 27.6 & 36.8 \\
& DGAM~\cite{dgam} (CVPR'20) & I3D & 60.0 & 54.2 & 46.8 & 38.2 & 28.8 & 19.8 & 11.4 & 45.6 & 29.0 & 37.0 \\
& TSCN~\cite{TSCN} (ECCV'20) & I3D & 63.4 & 57.6 & 47.8 & 37.7 & 28.7 & 19.4 & 10.2 & 47.0 & 28.8 & 37.8 \\
& EM-MIL~\cite{luo2020emmil} (ECCV'20) & I3D & 59.1 & 52.7 & 45.5 & 36.8 & 30.5 & 22.7 & \textbf{16.4} & 45.0 & 30.4 & 37.7 \\
& A2CL-PT~\cite{min2020adversarial} (ECCV'20) & I3D & 61.2 & 56.1 & 48.1 & 39.0 & 30.1 & 19.2 & 10.6 & 46.9 & 29.4 & 37.8 \\
& HAM-Net~\cite{islam2021hybrid} (AAAI'21) & I3D & 65.4 & 59.0 & 50.3 & 41.1 & 31.0 & 20.7 & 11.1 & 49.4 & 30.8 & 39.8 \\
& SUB-ACT~\cite{huangModelingSubActionsWeakly2021} (TCSVT'22) & I3D & 66.1 & 60.0 & 52.3 & 43.2 & 32.9 & - & - & 50.9 & - & - \\
& AUMN~\cite{luoActionUnitMemory2021} (CVPR'21) & I3D & 66.2 & 61.9 & 54.9 & 44.4 & 33.3 & 20.5 & 9.0 & 52.1 & 32.4 & 41.5 \\
& CoLA~\cite{zhangCoLAWeaklySupervisedTemporal2021} (CVPR'21) & I3D & 66.2 & 59.5 & 51.5 & 41.9 & 32.2 & 22.0 & 13.1 & 50.3 & 32.1 & 40.9 \\
& UGCT~\cite{yangUncertaintyGuidedCollaborative2021} (CVPR'21) & I3D & 69.2 & 62.9 & 55.5 & 46.5 & 35.9 & 23.8 & 11.4 & 54.0 & 34.6 & 43.6 \\
& D2-Net~\cite{narayan2021d2} (ICCV'21) & I3D & 65.7 & 60.2 & 52.3 & 43.4 & 36.0 & - & - & 51.5 & - & - \\
& FAC-Net~\cite{huangForegroundActionConsistencyNetwork2021} (ICCV'21) & I3D & 67.6 & 62.1 & 52.6 & 44.3 & 33.4 & 22.5 & 12.7 & 52.0 & 33.1 & 42.2 \\
& DCC~\cite{li2022dcc} (CVPR'22) & I3D & 69.0 & 63.8 & 55.9 & 45.9 & 35.7 & 24.3 & 13.7 & 54.1 & 35.1 & 44.0 \\
& ASM-Loc~\cite{heASMLocActionawareSegment2022} (CVPR'22) & I3D & 71.2 & 65.5 & 57.1 & 46.8 & 36.6 & 25.2 & 13.4 & 55.4 & 35.8 & 45.1 \\
& Huang \textit{el al.}~\cite{huang2022weakly} (CVPR'22) & I3D & 71.3 & 65.3 & 55.8 & 47.5 & 38.2 & 25.4 & 12.5 & 55.6 & 35.9 & 45.1 \\
\cline{2-13}
& \textbf{Ours} & I3D & \textbf{72.6} & \textbf{67.1} & \textbf{59.5} & \textbf{49.3} & \textbf{39.4} & \textbf{26.5} & 13.4 & \textbf{57.2} & \textbf{37.0} & \textbf{46.8} \\
\hline
\hline

\end{tabular}}
\vspace{-4mm}
\end{center}
\end{table*}

\begin{table}[!t]
\begin{center}
\caption{Comparisons of temporal localization on ActivityNet1.3 Dataset. AVG indicates the Average mAP at IoU thresholds ranging from 0.5 to 0.95 with an interval of 0.05. }
\label{table:activity}
\begin{tabular}{c|l|ccc|c}
\hline
\hline
\multirow{2}{*}{Supervision} & \multirow{2}{*}{Method} & \multicolumn{3}{c|}{mAP @ IoU (\%)} & \multirow{2}{*}{AVG} \\
\cline{3-5}
&  & 0.5 & 0.75 & 0.95 &  \\
\hline
\hline
\multirow{4}{*}{{Full}}
& {BSN~\cite{lin2018bsn}} & 46.45 & 29.96 & 8.02 & 30.03 \\	
& {BMN~\cite{lin2019bmn}} & 50.36 & 34.60	& 9.02 & 34.09 \\
& {G-TAD~\cite{xu2020gtad}} & 50.07 & 34.78 & 8.29 & 33.85 \\
& {TadTR~\cite{TadTR}} & 53.62 & 37.52 & 10.56 & 36.75  \\		
\hline
\multirow{10}{*}{{Weak}}
& STPN~\cite{stpn} & 29.3 & 16.9 & 2.6 & 16.3 \\
& CMCS~\cite{hongCrossmodalConsensusNetwork2021} & 34.0 & 20.9 & 5.7 & 21.2 \\
& BasNet~\cite{leeBackgroundSuppressionNetwork2020} & 34.5 & 22.5 & 4.9 & 22.2 \\
& UGCT~\cite{yangUncertaintyGuidedCollaborative2021} & 39.1 & 22.4 & 5.8 & 23.8 \\
& FAC-Net~\cite{huangForegroundActionConsistencyNetwork2021} & 37.6 & 24.2 & 6.0 & 24.0 \\
& FTCL~\cite{gao2022fine} & 40.0 & 24.3 & 6.4 & 24.8 \\
& DCC~\cite{liExploringDenoisedCrossvideo2022} & 38.8 & 24.2 & 5.7 & 24.3 \\
& Huang \textit{et al.}~\cite{huang2022weakly} & 40.6 & 24.6 & 5.9 & 25.0 \\
& ASM-Loc~\cite{heASMLocActionawareSegment2022} & 41.0 & 24.9 & 6.2 & 25.1 \\
\cline{2-6}
& \textbf{Ours} & \textbf{41.6} & \textbf{25.1} & \textbf{6.5} & \textbf{25.3} \\
\hline
\hline
\end{tabular}
\vspace{-4mm}
\end{center}
\end{table}
\begin{table}[!t]
\begin{center}
\caption{Comparisons of temporal localization on FineAction Dataset. AVG indicates the Average mAP at IoU thresholds ranging from 0.5 to 0.95 with an interval of 0.05.}
\label{table:fineaction}
\begin{tabular}{c|l|ccc|c}
\hline
\hline
\multirow{2}{*}{Supervision} & \multirow{2}{*}{Method} & \multicolumn{3}{c|}{mAP @ IoU (\%)} & \multirow{2}{*}{AVG} \\
\cline{3-5}
&  & 0.5 & 0.75 & 0.95 &  \\
\hline
\hline
\multirow{4}{*}{{Full}}
& {BSN~\cite{lin2018bsn}} & 10.65 & 6.43	& 2.50 & 6.75\\
& {BMN~\cite{lin2019bmn}} & 14.44 & 8.92 & 3.12 & 9.25 \\
& {G-TAD~\cite{xu2020gtad}} & 13.74 & 8.83 & 3.06 & 9.06\\
& {TadTR~\cite{lin2019bmn}} & 29.07	& 17.66	& 5.07 & 18.24 \\
\hline
\multirow{6}{*}{{Weak}}
& W-TALC~\cite{paulWTALCWeaklySupervisedTemporal2018} & 6.18 & 3.15 & 0.83 & 3.45 \\
& BasNet~\cite{leeBackgroundSuppressionNetwork2020} & 6.65 & 3.23 & 0.95 & 3.64 \\
& ASL~\cite{ASL} & 6.79 & 2.68 & 0.81 & 3.30 \\
& D2-Net~\cite{narayan2021d2} & 6.75 & 3.02 & 0.82 & 3.35 \\
& HAAN~\cite{HAAN} & 7.05 & 3.95 & 1.14 & 4.10 \\
\cline{2-6}
& \textbf{Ours} & \textbf{7.69} & \textbf{4.81} & \textbf{1.32} & \textbf{4.58} \\
\hline
\hline
\end{tabular}
\vspace{-4mm}
\end{center}
\end{table}

\subsubsection{\textbf{ActivityNet Dataset}}
ActivityNet~\cite{caba2015activitynet} is a benchmark dataset for temporal action localization tasks and is available in two versions, i.e., 1.2 and 1.3.
In this work, we utilize the training set of version 1.3.
ActivityNet 1.3 includes 200 complex human action classes and provides 10,024 videos for training, 4,926 videos for validation, and 5,044 videos for testing.
We use the training set to train our model and the validation set to evaluate the performance of our model.
\par

\subsubsection{\textbf{FineAction Dataset}}

FineAction~\cite{liu2022fineaction} combines three existing datasets. YouTube8M, Kinetics400, and FCVID.
This dataset contains a wide range of video content with a three-level tagging hierarchy from coarse to fine.
In this work, we used the same setup as~\cite{HAAN} and trained the model using labels from the middle level.

\subsubsection{\textbf{Evaluation Metric}}
Following previous work~\cite{luoActionUnitMemory2021,quACMNetActionContext2021,heASMLocActionawareSegment2022}, we evaluate our methods on three benchmark datasets with mean Average Precision (mAP) under different t-IoU thresholds. The t-IoU thresholds for THUMOS14 are [0.1:0.1:0.7],  for ActivityNet1.3 is [0.5:0.05:0.95], and for FineAction is [0.5:0.05:0.95].

\subsection{Implementation Details}
In this section, we provide details on the model parameters and training parameters.

\subsubsection{\textbf{Model Details}}
The model network is implemented in Pytorch using two Nvidia GeForce 2080Ti GPUs.
Following previous methods~\cite{heASMLocActionawareSegment2022, leeBackgroundSuppressionNetwork2020}, we use I3D pre-trained on the Kinetics dataset to extract the video feature without fine-tuning.
The dimension of the extracted feature is 2048.
The number of sampling segments $T$ for THUMOS14, ActivityNet, and FineAction is set to 750, 75, and 400, respectively.
The memory queue length of each action class is set to $500$ to store the discriminative action features at the dataset level.
The knowledge summarization and aggregation modules are implemented according to~\cite{attention}.
The attention head and classification head are implemented by a set of MLP layers.
The number of learnable latent code words is set to $40$ and $200$ for THUMOS14 and ActivityNet, respectively.

\subsubsection{\textbf{Training Details}}
For datasets THUMOS14, ActivityNet1.3, and FineAction, we set the batch sizes to 16, 64, and 32, respectively, and use the ADAM~\cite{adam} optimizer to train our model.
The number of total training epochs is fixed at 200.
The initial learning rate is set to $1e^{-4}$ and decays by 0.1 for every 100 epochs.
{Since representative segments are not well learned in the early stages of training, storing them directly in the memory bank can introduce additional noise and make the whole training process unstable. 
Therefore, to ensure the stability of the training process, we build the memory bank for feature storage and cross-video information interaction only after 50 training epochs.}
According to the hyper-parameter setting, in the memory-guided contraction learning section, we set the momentum coefficient $\alpha$ to 0.99. The parameters $\lambda$ and $q$ in the robust contrast loss are set to 0.5 and 0.2, respectively.
In the objective function, the coefficient of the variance regularization term $\rho$ is set to 0.001, and the parameters $\mu$ and $\gamma$, which are used to balance different losses, are set to 0.1 and 1, respectively.
Noted that, our method only utilizes the pre-trained I3D model~\cite{I3D} to extract video features. 
Apart from the feature extraction module, all other modules of the proposed model are trained from scratch on the training dataset.
Besides, to ensure a fair comparison with other methods, we refrain from performing any additional fine-tuning operations on the pre-trained I3D backbone.

\subsection{Comparison with State-of-the-Art Methods}
In this section, we compared our approach with the current state-of-the-art temporal action localization methods, both weakly supervised and fully supervised.
It can be observed that our weakly supervised localization method can achieve comparable results to the recent fully supervised methods on the THUMO14 dataset. However, there is still a performance gap between our method and fully supervised methods on more challenging datasets such as ActivityNet 1.3 and FineAction.

\subsubsection{\textbf{Evaluation on THUMOS14}}
As shown in Table~\ref{table:THUMOS14}, our method outperforms previous methods in almost all metrics on the THUMOS14 dataset. Compared to previous methods, our method outperforms the state-of-the-art method by $1.8\%$.
This demonstrates the effectiveness of our proposed framework and implies that our method produces more accurate and complete action localizations.
Furthermore, although our method is trained using weak labels, our method is on par with the performance of several fully-supervised methods and can outperform some semi-supervised methods that use additional label information during the training process.
\par

\subsubsection{\textbf{Evaluation on ActivityNet1.3}}
As shown in Table~\ref{table:activity}, our method achieves the best performance on the Activity1.3 dataset. The performance of our method outperforms the previous best method~\cite{heASMLocActionawareSegment2022} by $1.4\%$.
Noted that \cite{heASMLocActionawareSegment2022} uses intra- and inter-segment attention to improve the modeling of action-aware segments.
However, our method combines robust contrast learning with global knowledge aggregation to constrain the cross-video consistency of action segment features.
Due to the large number of short actions that exist in the ActivityNet1.3 dataset, the improvement of our method on this dataset is not as significant as the improvement on the THUMOS14 dataset.

\subsubsection{\textbf{Evaluation on FineAction}}
By using the open source code of previous methods, we can reproduce the performance of baseline WSTAL methods on the FineAction dataset.
As shown in Table~\ref{table:fineaction}, we compare our method with the reproduced results, it can be observed that our methods outperform previous methods by a large margin.
The regional contrastive learning strategy allows for the separation of features from different action classes in feature space, which improves the accuracy of fine-grained action classification and localization.

\begin{table}[!t]
\begin{center}
\caption{Ablation studies for each of the proposed components.
RMGCL in the table denotes robust contrastive learning strategy, \textit{GKS} denotes feature summarization module, \textit{GKA} denotes the regional feature propagation/aggregation, and \textit{Pseudo} indicates pseudo-label-based learning.}
\label{table:ablation_component}
\begin{tabular}{cccc|ccc}
\hline \hline
\multicolumn{4}{c|}{\textbf{Method}}                      & \multicolumn{3}{c}{\textbf{mAP @ IoU}} \\ \hline 
RMGCL     & GKS    & GKA  & Pseudo & 0.1         & 0.3         & AVG        \\ \hline \hline
             &              &              &              & 68.9        & 55.0        & 42.6       \\
$\checkmark$ &              &              &              & 69.1        & 55.3        & 43.2       \\
$\checkmark$ &              & $\checkmark$ &              & 67.5        & 53.2        & 42.4       \\
$\checkmark$ &              & $\checkmark$ & $\checkmark$ & 67.3        & 53.0        & 42.1       \\
$\checkmark$ & $\checkmark$ & $\checkmark$ &              & 71.5        & 56.1        & 45.3       \\
$\checkmark$ & $\checkmark$ & $\checkmark$ & $\checkmark$ & \textbf{72.6}        & \textbf{59.4}        & \textbf{46.8}       \\ \hline \hline
\end{tabular}
\vspace{-4mm}
\end{center}
\end{table}

\begin{table}[!t]
\begin{center}
\caption{Ablation studies of contrastive learning strategy. \textit{Baseline} model we use a model similar to ACM-Net~\cite{quACMNetActionContext2021}.}
\label{table:contrastive}
\begin{tabular}{l|ccc}
\hline \hline
\multirow{2}{*}{\textbf{Method}} & \multicolumn{3}{c}{\textbf{mAP @ IoU}} \\ \cline{2-4} 
                                 & 0.1         & 0.3         & AVG        \\ \hline \hline
Baseline                         & 68.9        & 55.0        & 42.6       \\ \hline
+ dataset-level memory bank              & 69.1        & 55.3        & 42.3       \\
+ background          & 69.8        & 56.4        & 43.7       \\
+ robust contrastive loss       & \textbf{70.3}        & \textbf{57.1}        & \textbf{44.8}       \\ \hline \hline
\end{tabular}
\vspace{-4mm}
\end{center}
\end{table}

\subsection{Ablation Studies}
\label{sec:ablation_study}
In this section, we performed a series of ablation studies on THUMOS14 to demonstrate the effectiveness of our method.
\par

\subsubsection{\textbf{Contribution of each Component}}
We evaluated the effectiveness of our proposed modules and the results are shown in Table~\ref{table:ablation_component}.
In our experimental results, we can observe that the adoption of regional feature comparison learning is effective in improving the performance of the model.
However, when we attempt to propagate the features from the memory bank directly to the current video segments, the model performance shows a slight drop.
Our analysis suggests that this is due to the fact that there are still large variations in the stored action features, which can introduce additional noise information into the current video features.
When we introduced the knowledge summarization module, our model achieved gains of $5.9\%$, proving the effectiveness of our proposed module.
After performing knowledge aggregation to obtain new features, the online supervision of the generated pseudo-labels can further improve the performance of the model by $1.5\%$.
The best performance metrics of the model can be achieved only when combining our proposed four modules.
\par
\begin{figure}[t!]
    \centering
    \includegraphics[width=0.4\textwidth]{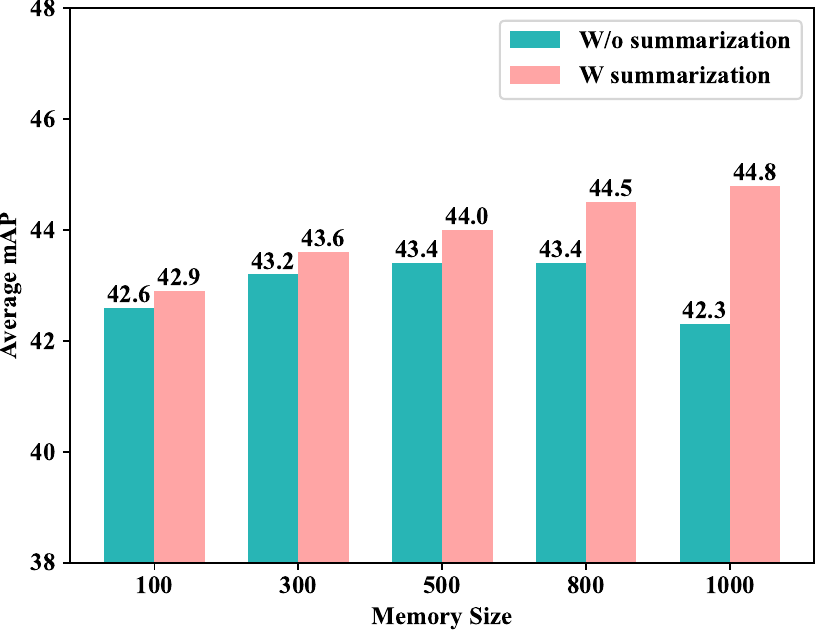}
    \caption{\textbf{Ablation studies of Memory size.} The number of stored features can affect the quality of global contextual knowledge and the quality of action localization.}
    \label{fig:memorysize}
    \vspace{-4mm}
\end{figure}
\begin{table}[!t]
\begin{center}
\caption{Comparision of different approaches of generating the summarization of action segments. We compare our latent codewords method with several clustering methods.}
\label{table:cluster_method}
\begin{tabular}{ll|ccc}
\hline \hline
\multicolumn{2}{c|}{\multirow{2}{*}{\textbf{Method}}} & \multicolumn{3}{c}{\textbf{mAP @ IoU}}        \\ \cline{3-5} 
\multicolumn{2}{c|}{}                                 & 0.1           & 0.3           & AVG           \\ \hline \hline
\multicolumn{1}{l|}{\multirow{4}{*}{\begin{tabular}[c]{@{}l@{}}Non-Learnable \\ Methods\end{tabular}}} & K-means & 71.2 & 57.1 & 45.4 \\
\multicolumn{1}{l|}{}      & Spectral Clustering      & 70.3          & 56.1          & 44.7          \\
\multicolumn{1}{l|}{}      & DBSCAN~\cite{DBSCAN}                  & 69.8          & 55.2          & 44.2          \\
\multicolumn{1}{l|}{}      & Optimal Transport~\cite{asano2019self}        & 70.8          & 56.6          & 45.3          \\ \hline
\multicolumn{1}{l|}{\multirow{2}{*}{\begin{tabular}[c]{@{}l@{}}Learnable\\ Method\end{tabular}}}       & EM-Attention~\cite{li2019expectation}  & 71.4 & 57.1 & 45.6 \\
\multicolumn{1}{l|}{}      & Latent codewords (ours)        & \textbf{72.5} & \textbf{59.4} & \textbf{46.7} \\ \hline \hline
\end{tabular}
\vspace{-4mm}
\end{center}
\end{table}

\subsubsection{\textbf{Memory-Guided Contrastive Learning}}
We first define a baseline model, which is the same as ACM-Net~\cite{quACMNetActionContext2021}.
The experiments in Table~\ref{table:contrastive} show that the performance of the model can be improved by introducing a dataset-level repository.
Compared with previous methods, this repository can store and integrate features from different videos, providing a holistic understanding of the action features. 
In addition, we attempted to store representative features of the background segments in the memory bank.
The background features are often noisy and can vary greatly between videos, making it challenging to integrate and provide a holistic understanding of the backgrounds.
However, experiments demonstrate that this strategy improves the performance of our model. 
We attribute this to the fact that, via our contrast learning strategy, the model learns a universal pattern of foreground and background features.

In Fig.~\ref{fig:memorysize}, we evaluate the effect of different numbers of stored features on the performance of the model.
The results show that increasing the number of stored features can improve the performance of the model in a short term.
However, as the number of stored features increases, it leads to a performance bottleneck in the model due to the large variance of the features, which can introduce additional noise into the model.
This problem is alleviated by introducing the knowledge summarization module, which obtains a more compact feature representation while smoothing out the noise in the memory bank.
\par

\begin{figure}[t!]
    \centering
    \includegraphics[width=0.48\textwidth]{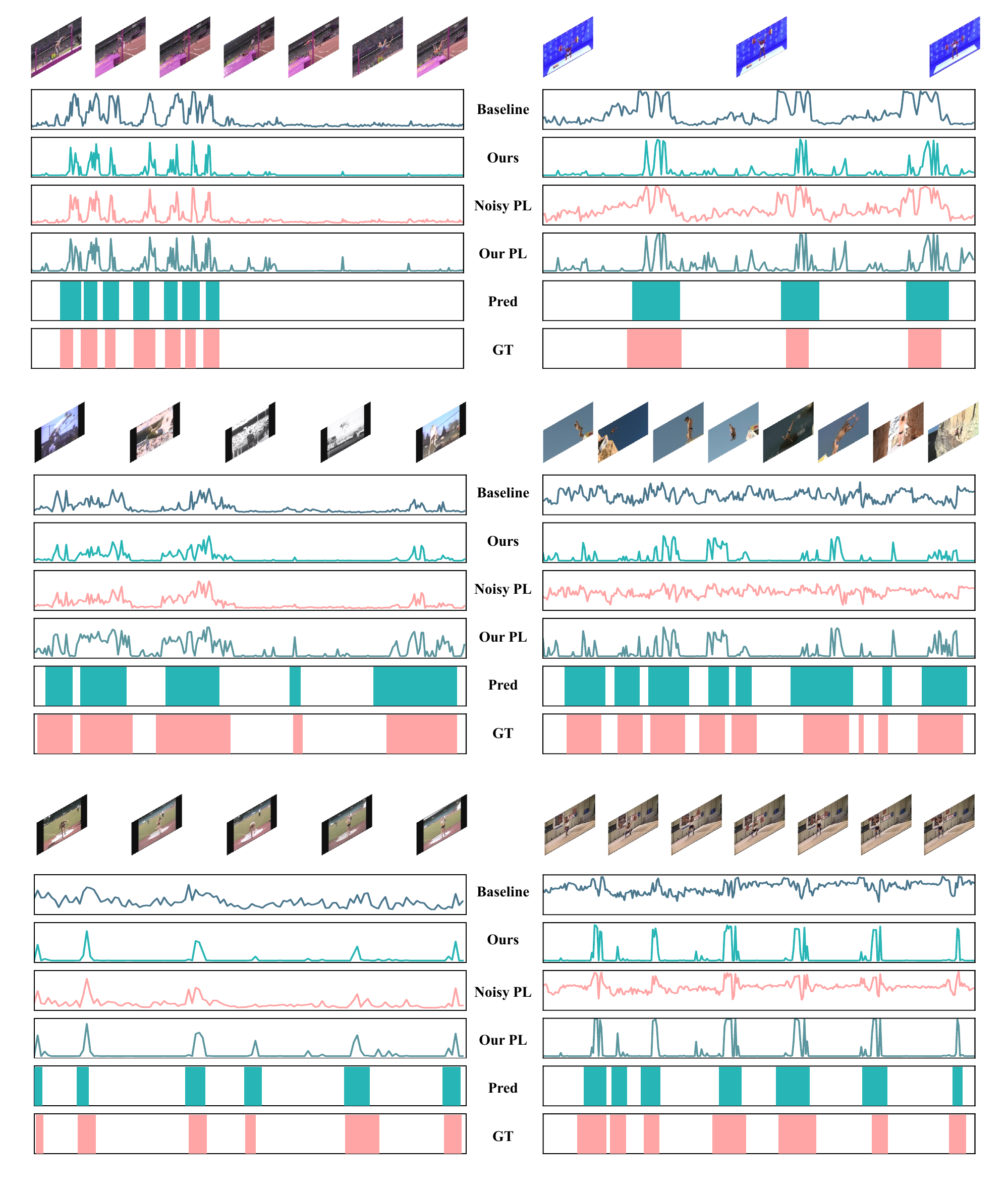}
    \caption{{\textbf{Qualitative results.} 
    We show the qualitative results on THUMOS14~\cite{idreesTHUMOSChallengeAction2017}. PL denotes the pseudo label.
    Noisy PL is obtained by TSCN~\cite{TSCN}.
    With the proposed module, our method enhances the holistic understanding of action patterns, which in turn generates higher-quality pseudo-labels providing frame-level supervision to the main branch and alleviating localization ambiguity.}}
    \label{fig:qualitative}
    \vspace{-4mm}
\end{figure}
\begin{figure}[t!]
    \centering
    \includegraphics[width=0.40\textwidth]{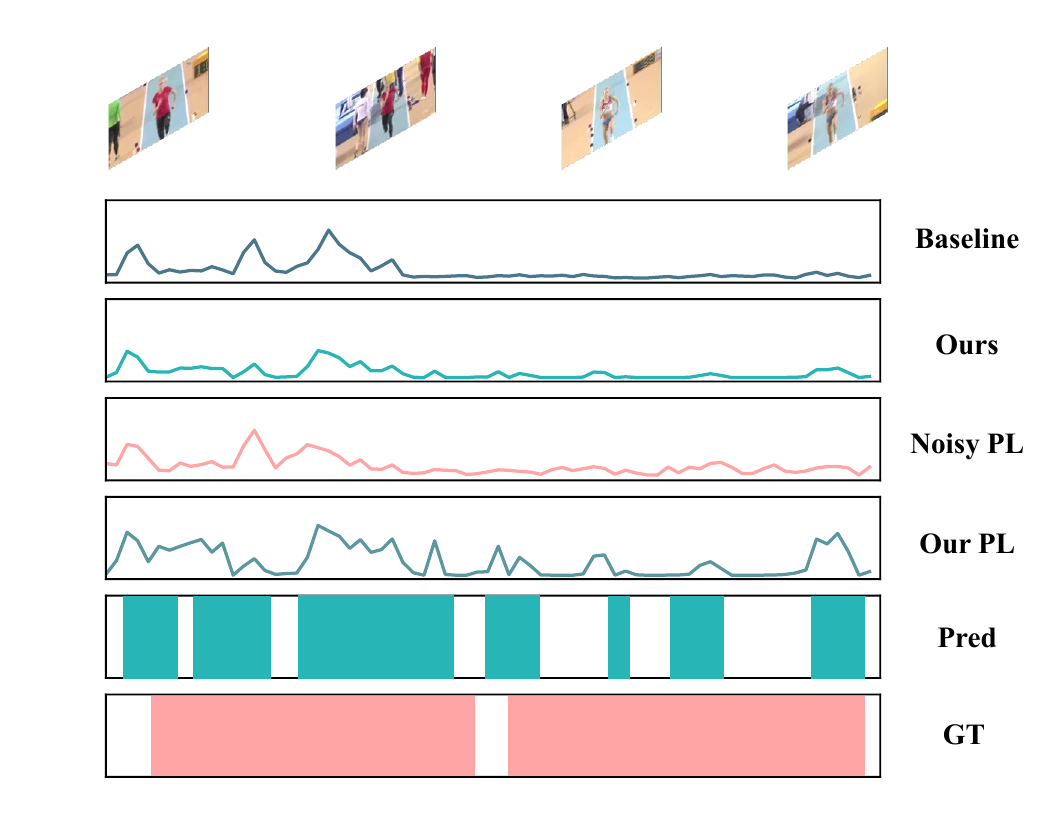}
    \caption{{\textbf{Failure case.} We show some failures of our approach. In our experiments, we found that our method has difficulty in capturing complete action instances with slow-motion features~\cite{sunSlowMotionMatters2022}.}}
    \label{fig:failure}
    \vspace{-4mm}
\end{figure}

\begin{table}[!t]
\begin{center}
\caption{Ablation studies for the number of latent codewords. }
\label{table:latent_num}
\begin{tabular}{c|clll}
\hline \hline
\multirow{2}{*}{\textbf{Method}} & \multicolumn{4}{c}{\textbf{mAP @ IoU}}                                                      \\ \cline{2-5} 
                                 & 0.1           & \multicolumn{1}{c}{0.3} & \multicolumn{1}{c}{0.5} & \multicolumn{1}{c}{AVG} \\ \hline \hline
10                               & 71.5          & 57.3                    & 38.4                    & 45.8                    \\
20                               & 72.3          & 58.0                    & 39.2                    & 46.6                    \\
\textbf{40}                      & \textbf{72.6} & \textbf{59.4}           & \textbf{39.3}           & \textbf{46.7}           \\
60                               & 72.5          & 58.2                    & 39.4                    & 46.7                    \\
80                               & 72.0          & 57.9                    & 38.9                    & 46.3                    \\
100                              & 72.1          & 58.0                    & 39.0                    & 46.4                    \\ \hline \hline
\end{tabular}
\vspace{-4mm}
\end{center}
\end{table}

\begin{table}[!t]
\begin{center}
\caption{Evaluation of the generalized performance of our proposed module on different action location approaches.}
\label{table:generalized_evaluation}
\begin{tabular}{l|cccc}
\hline \hline
\multirow{2}{*}{\textbf{Method}} & \multicolumn{4}{c}{\textbf{mAP @ IoU}}                        \\ \cline{2-5} 
                                 & 0.1           & 0.3           & 0.5           & AVG           \\ \hline \hline
BasNet~\cite{leeBackgroundSuppressionNetwork2020}                           & 61.7          & 48.2          & 29.3          & 37.7          \\
BasNet + ours                    & \textbf{69.2} & \textbf{54.7} & \textbf{36.1} & \textbf{43.8} \\ \hline
FAC-Net~\cite{huangForegroundActionConsistencyNetwork2021}                          & 67.6          & 52.6          & 33.4          & 42.2          \\
FAC-Net + ours                   & \textbf{71.4} & \textbf{57.2} & \textbf{38.3} & \textbf{45.8} \\ \hline
STPN~\cite{stpn}                             & 57.0          & 42.8          & 24.7          & 33.5          \\
STPN + ours                      & \textbf{66.4} & \textbf{52.0} & \textbf{33.2} & \textbf{41.4} \\ \hline \hline
\end{tabular}
\vspace{-7mm}
\end{center}
\end{table}

\subsubsection{\textbf{Knowledge Summarization}}
We explored a variety of methods for performing knowledge summarization, and the experiment results are shown in Table~\ref{table:cluster_method}.
We mainly compared some non-learning methods based on iterative optimization and some learning-based methods.
The results show that our methods achieve the best performance.
Noted that traditional clustering methods have many parameters that need to be manually adjusted, which has a large impact on final performance.
In addition, the iterative process cannot be optimized in parallel, resulting in a longer training time.

In order to investigate the effect of the number of words on the performance of the model, we conducted a comparison experiment, and the results are shown in Table~\ref{table:latent_num}.
It can be observed that adjusting the number of latent codewords can improve the performance of the model.
However, the model is not sensitive to this parameter and does not need to be adjusted manually, and the model achieves a performance improvement under most of the parameter settings.
\par

\begin{table}[!t]
\begin{center}
\caption{Computation Complexity of proposed method. }
\label{table:computation}
\begin{tabular}{c|c|ccc}
\hline \hline
\textbf{Model} & \textbf{\begin{tabular}[c]{@{}c@{}}Input\\ dimension\end{tabular}} & \textbf{Params(M)} & \textbf{FLOPs(G)} & \textbf{\begin{tabular}[c]{@{}c@{}}Model\\ Size(MB)\end{tabular}} \\ \hline \hline
\multirow{3}{*}{\begin{tabular}[c]{@{}c@{}}Training\\ Stage\end{tabular}} & {[}1, 750, 2048{]} & 84.16 & 9.453 & 106.01 \\
 & {[}1, 75, 2048{]} & 84.16 & 0.952 & 89.29 \\
 & {[}1, 400, 2048{]} & 84.16 & 5.045 & 97.34 \\ \hline
\multirow{3}{*}{\begin{tabular}[c]{@{}c@{}}Testing\\ Stage\end{tabular}} & {[}1, 750, 2048{]} & 50.54 & 8.679 & 69.11 \\
 & {[}1, 75, 2048{]} & 50.54 & 0.944 & 52.39 \\
 & {[}1, 400, 2048{]} & 50.54 & 4.686 & 60.44 \\ \hline \hline
\end{tabular}%
\vspace{-4mm}
\end{center}
\end{table}


\subsubsection{\textbf{Generalize Ability}}
We evaluate the generalization ability of our method by applying our algorithm to three recent baseline models FAC-Net~\cite{huangForegroundActionConsistencyNetwork2021}, STPN~\cite{stpn} and BaSNet~\cite{leeBackgroundSuppressionNetwork2020}.
The experiment results are shown in Table~\ref{table:generalized_evaluation}.
The performance of these methods is improved significantly after integration with our method.
This proves that our method has a promising generalization ability on different methods.


\par

\subsection{\textbf{Computation Complexity Analysis}}
\label{sec:Complexity}
To assess the practicality of our proposed model, we conducted an evaluation of its complexity on a single Nvidia GeForce 2080Ti GPU using three key metrics: the total number of model parameters (Params), the number of floating-point operations (FLOPs), and the overall size of the model (including forward/backward pass sizes and parameter sizes).
In addition, we also computed these metrics under different input dimensions.
As shown in Table~\ref{table:computation}, during the training stage, our model exhibits higher memory usage and computational requirements due to the inclusion of a memory bank. This is attributed to the necessity of storing and updating the memory bank during training, which introduces additional parameters. Consequently, the total number of model parameters is high during the training phase.
However, during the testing stage, our model demonstrates computational efficiency and minimal memory consumption. This is because we solely utilize latent representations $T_M$ to aggregate cross-video features, and the memory bank is not utilized during this stage. As a result, the total number of model parameters is significantly reduced during testing.

\subsection{Qualitative Results.}
We visualize some examples of detected actions and generated pseudo-labels in Fig.~\ref{fig:qualitative}. 
Noted that we utilize TSCN~\cite{TSCN} to generate noisy pseudo-label (Noisy PL). 
TSCN adopts the idea of two-stream late fusion to generate pseudo-labels, which emphasize locations with high activations in both streams, indicating actual action instances. Low activations in both streams represent the background while high activations in only one stream may indicate false positives.
Unlike TSCN, our method can obtain high-confidence pseudo-labels by efficiently fusing global video contextual information as a way to further reduce localization ambiguities.
As can be observed from the results we provide, our method can generate more accurate pseudo-labels.
At the same time, by exploiting the knowledge of representative features, the noisy fragments in the background can be effectively suppressed, resulting in more accurate localization results.
However, as can be seen in Fig.~\ref{fig:failure}, the localization results of our method for actions with slow-motion features are not satisfactory, as they have similar characteristics and thus can be easily confused with background instances.
This problem can be alleviated by utilizing the method in~\cite{sunSlowMotionMatters2022}.

\section{CONCLUSION}

In this paper, we propose a robust cross-video contrast and aggregation learning framework, including the RMGCL module and the GKSA module, to address two core issues existing in recent WSTAL pipelines, i.e., ambiguous classification learning and localization ambiguity.
In order to alleviate the ambiguity caused by the limited available knowledge carried by video-level labels, our method explores the possibility of recovering holistic action semantic structure from the perspective of cross-video context mining.
Specifically, a continuously updated memory bank is first constructed to store global representative action knowledge.
The RMGCL helps the model learn a more structured and compact action embedding space with the assistance of the memory bank, which can enhance the comprehensive understanding of various fine-grained action patterns and thus alleviate the ambiguity in classification learning.
Subsequently, we use GKSA to efficiently gather cross-video representative knowledge and holistic contextual cues stored in the memory bank in a learnable manner to generate high-quality frame-level pseudo-labels for self-learning, assisting in mitigating localization ambiguity.
Our method improves average mAP by $1.8\%$, $1.4\%$, and $1.2\%$ on the THUMOS14, ActivityNet1.3, and FineAction datasets, respectively, compared to the previous SOTA methods, and can be easily plugged into the existing WSTAL methods bringing performance improvements.
One potential limitation of our method is the need to maintain a large action memory bank during training and perform cross-video contrastive learning to eliminate the ambiguities in the pipeline.
This results in our model consuming a large amount of GPU memory and a long training time. Additionally, significant domain gaps between the training and testing phases could potentially impact the learning of the memory bank, which in turn could adversely affect the quality of the generated pseudo-labels.

\bibliographystyle{IEEEtran}
\bibliography{IEEEabrv,ref}

\begin{thebibliography}{10}
\providecommand{\url}[1]{#1}
\csname url@samestyle\endcsname
\providecommand{\newblock}{\relax}
\providecommand{\bibinfo}[2]{#2}
\providecommand{\BIBentrySTDinterwordspacing}{\spaceskip=0pt\relax}
\providecommand{\BIBentryALTinterwordstretchfactor}{4}
\providecommand{\BIBentryALTinterwordspacing}{\spaceskip=\fontdimen2\font plus
\BIBentryALTinterwordstretchfactor\fontdimen3\font minus \fontdimen4\font\relax}
\providecommand{\BIBforeignlanguage}[2]{{%
\expandafter\ifx\csname l@#1\endcsname\relax
\typeout{** WARNING: IEEEtran.bst: No hyphenation pattern has been}%
\typeout{** loaded for the language `#1'. Using the pattern for}%
\typeout{** the default language instead.}%
\else
\language=\csname l@#1\endcsname
\fi
#2}}
\providecommand{\BIBdecl}{\relax}
\BIBdecl

\bibitem{zhou2016learning}
B.~Zhou, A.~Khosla, A.~Lapedriza, A.~Oliva, and A.~Torralba, ``Learning deep features for discriminative localization,'' in \emph{Proceedings of the IEEE conference on computer vision and pattern recognition}, 2016, pp. 2921--2929.

\bibitem{huang2021foreground}
L.~Huang, L.~Wang, and H.~Li, ``Foreground-action consistency network for weakly supervised temporal action localization,'' in \emph{Proceedings of the IEEE/CVF International Conference on Computer Vision}, 2021, pp. 8002--8011.

\bibitem{huang2022weakly}
L.~\vspace{0mm}Huang, L.~Wang, and H.~Li, ``Weakly supervised temporal action localization via representative snippet knowledge propagation,'' in \emph{Proceedings of the IEEE/CVF Conference on Computer Vision and Pattern Recognition}, 2022, pp. 3272--3281.

\bibitem{leeBackgroundSuppressionNetwork2020}
P.~Lee, Y.~Uh, and H.~Byun, ``Background {{Suppression Network}} for {{Weakly-Supervised Temporal Action Localization}},'' in \emph{Proceedings of the {{AAAI Conference}} on {{Artificial Intelligence}}}, vol.~34, Apr. 2020, pp. 11\,320--11\,327.

\bibitem{stpn}
P.~Nguyen, T.~Liu, G.~Prasad, and B.~Han, ``Weakly supervised action localization by sparse temporal pooling network,'' in \emph{Proceedings of the IEEE Conference on Computer Vision and Pattern Recognition}, 2018, pp. 6752--6761.

\bibitem{feng2022bias}
L.~Feng, C.~Zhao, and X.~Li, ``Bias-eliminated semantic refinement for any-shot learning,'' \emph{IEEE Transactions on Image Processing}, vol.~31, pp. 2229--2244, 2022.

\bibitem{gan2015devnet}
C.~Gan, N.~Wang, Y.~Yang, D.-Y. Yeung, and A.~G. Hauptmann, ``Devnet: A deep event network for multimedia event detection and evidence recounting,'' in \emph{Proceedings of the IEEE Conference on Computer Vision and Pattern Recognition}, 2015, pp. 2568--2577.

\bibitem{huangRelationalPrototypicalNetwork2020}
L.~Huang, Y.~Huang, W.~Ouyang, and L.~Wang, ``Relational {{Prototypical Network}} for {{Weakly Supervised Temporal Action Localization}},'' in \emph{Proceedings of the {{AAAI Conference}} on {{Artificial Intelligence}}}, vol.~34, Apr. 2020, pp. 11\,053--11\,060.

\bibitem{zhangCoLAWeaklySupervisedTemporal2021}
C.~Zhang, M.~Cao, D.~Yang, J.~Chen, and Y.~Zou, ``{{CoLA}}: {{Weakly-Supervised Temporal Action Localization}} with {{Snippet Contrastive Learning}},'' in \emph{2021 {{IEEE}}/{{CVF Conference}} on {{Computer Vision}} and {{Pattern Recognition}} ({{CVPR}})}.\hskip 1em plus 0.5em minus 0.4em\relax {Nashville, TN, USA}: {IEEE}, Jun. 2021, pp. 16\,005--16\,014.

\bibitem{li2022dcc}
J.~Li, T.~Yang, W.~Ji, J.~Wang, and L.~Cheng, ``Exploring denoised cross-video contrast for weakly-supervised temporal action localization,'' in \emph{Proceedings of the IEEE/CVF Conference on Computer Vision and Pattern Recognition}, 2022, pp. 19\,914--19\,924.

\bibitem{zhou2022object}
W.~Zhou, Y.~Li, and C.~Zhao, ``Object-guided and motion-refined attention network for video anomaly detection,'' in \emph{2022 IEEE International Conference on Multimedia and Expo (ICME)}.\hskip 1em plus 0.5em minus 0.4em\relax IEEE, 2022, pp. 1--6.

\bibitem{islamHybridAttentionMechanism2021}
A.~Islam, C.~Long, and R.~Radke, ``A {{Hybrid Attention Mechanism}} for {{Weakly-Supervised Temporal Action Localization}},'' in \emph{Proceedings of the {{AAAI Conference}} on {{Artificial Intelligence}}}, vol.~35, May 2021, pp. 1637--1645.

\bibitem{yangUncertaintyGuidedCollaborative2021}
W.~Yang, T.~Zhang, X.~Yu, T.~Qi, Y.~Zhang, and {FengWu}, ``Uncertainty {{Guided Collaborative Training}} for {{Weakly Supervised Temporal Action Detection}},'' in \emph{2021 {{IEEE}}/{{CVF Conference}} on {{Computer Vision}} and {{Pattern Recognition}} ({{CVPR}})}, Jun. 2021, pp. 53--63.

\bibitem{quACMNetActionContext2021}
S.~Qu, G.~Chen, Z.~Li, L.~Zhang, F.~Lu, and A.~Knoll, ``Acm-net: Action context modeling network for weakly-supervised temporal action localization,'' \emph{arXiv preprint arXiv:2104.02967}, 2021.

\bibitem{huangForegroundActionConsistencyNetwork2021}
L.~Huang, L.~Wang, and H.~Li, ``Foreground-{{Action Consistency Network}} for {{Weakly Supervised Temporal Action Localization}},'' in \emph{2021 {{IEEE}}/{{CVF International Conference}} on {{Computer Vision}} ({{ICCV}})}.\hskip 1em plus 0.5em minus 0.4em\relax {Montreal, QC, Canada}: {IEEE}, Oct. 2021, pp. 7982--7991.

\bibitem{wang2023two}
Y.~Wang, Y.~Li, and H.~Wang, ``Two-stream networks for weakly-supervised temporal action localization with semantic-aware mechanisms,'' in \emph{Proceedings of the IEEE/CVF Conference on Computer Vision and Pattern Recognition}, 2023, pp. 18\,878--18\,887.

\bibitem{zeng2019graph}
R.~Zeng, W.~Huang, M.~Tan, Y.~Rong, P.~Zhao, J.~Huang, and C.~Gan, ``Graph convolutional networks for temporal action localization,'' in \emph{Proceedings of the IEEE/CVF international conference on computer vision}, 2019, pp. 7094--7103.

\bibitem{luo2020emmil}
Z.~Luo, D.~Guillory, B.~Shi, W.~Ke, F.~Wan, T.~Darrell, and H.~Xu, ``Weakly-supervised action localization with expectation-maximization multi-instance learning,'' in \emph{Computer Vision--ECCV 2020: 16th European Conference, Glasgow, UK, August 23--28, 2020, Proceedings, Part XXIX 16}.\hskip 1em plus 0.5em minus 0.4em\relax Springer, 2020, pp. 729--745.

\bibitem{zhaiTwoStreamConsensusNetwork2020}
Y.~Zhai, L.~Wang, W.~Tang, Q.~Zhang, J.~Yuan, and G.~Hua, ``Two-{{Stream Consensus Network}} for~{{Weakly-Supervised Temporal Action~Localization}},'' in \emph{Computer {{Vision}} \textendash{} {{ECCV}} 2020}, ser. Lecture {{Notes}} in {{Computer Science}}, A.~Vedaldi, H.~Bischof, T.~Brox, and J.-M. Frahm, Eds.\hskip 1em plus 0.5em minus 0.4em\relax {Cham}: {Springer International Publishing}, 2020, pp. 37--54.

\bibitem{shouAutoLocWeaklySupervisedTemporal2018}
Z.~Shou, H.~Gao, L.~Zhang, K.~Miyazawa, and S.-F. Chang, ``{{AutoLoc}}: {{Weakly-Supervised Temporal Action Localization}} in {{Untrimmed Videos}},'' in \emph{Computer {{Vision}} \textendash{} {{ECCV}} 2018}, V.~Ferrari, M.~Hebert, C.~Sminchisescu, and Y.~Weiss, Eds., vol. 11220.\hskip 1em plus 0.5em minus 0.4em\relax {Cham}: {Springer International Publishing}, 2018, pp. 162--179.

\bibitem{heASMLocActionawareSegment2022}
B.~He, X.~Yang, L.~Kang, Z.~Cheng, X.~Zhou, and A.~Shrivastava, ``{{ASM-Loc}}: {{Action-aware Segment Modeling}} for {{Weakly-Supervised Temporal Action Localization}},'' in \emph{2022 {{IEEE}}/{{CVF Conference}} on {{Computer Vision}} and {{Pattern Recognition}} ({{CVPR}})}.\hskip 1em plus 0.5em minus 0.4em\relax {New Orleans, LA, USA}: {IEEE}, Jun. 2022, pp. 13\,915--13\,925.

\bibitem{ju2023distilling}
C.~Ju, K.~Zheng, J.~Liu, P.~Zhao, Y.~Zhang, J.~Chang, Q.~Tian, and Y.~Wang, ``Distilling vision-language pre-training to collaborate with weakly-supervised temporal action localization,'' in \emph{Proceedings of the IEEE/CVF Conference on Computer Vision and Pattern Recognition}, 2023, pp. 14\,751--14\,762.

\bibitem{idreesTHUMOSChallengeAction2017}
H.~Idrees, A.~R. Zamir, Y.-G. Jiang, A.~Gorban, I.~Laptev, R.~Sukthankar, and M.~Shah, ``The {{THUMOS}} challenge on action recognition for videos ``in the wild'','' \emph{Computer Vision and Image Understanding}, vol. 155, pp. 1--23, 2017.

\bibitem{caba2015activitynet}
F.~Caba~Heilbron, V.~Escorcia, B.~Ghanem, and J.~Carlos~Niebles, ``Activitynet: A large-scale video benchmark for human activity understanding,'' in \emph{Proceedings of the ieee conference on computer vision and pattern recognition}, 2015, pp. 961--970.

\bibitem{liu2022fineaction}
Y.~Liu, L.~Wang, Y.~Wang, X.~Ma, and Y.~Qiao, ``Fineaction: A fine-grained video dataset for temporal action localization,'' \emph{IEEE Transactions on Image Processing}, vol.~31, pp. 6937--6950, 2022.

\bibitem{lin2018bsn}
T.~Lin, X.~Zhao, H.~Su, C.~Wang, and M.~Yang, ``Bsn: Boundary sensitive network for temporal action proposal generation,'' in \emph{Proceedings of the European conference on computer vision (ECCV)}, 2018, pp. 3--19.

\bibitem{ssn}
Y.~Zhao, Y.~Xiong, L.~Wang, Z.~Wu, X.~Tang, and D.~Lin, ``Temporal action detection with structured segment networks,'' in \emph{Proceedings of the IEEE international conference on computer vision}, 2017, pp. 2914--2923.

\bibitem{wang2017untrimmednets}
L.~Wang, Y.~Xiong, D.~Lin, and L.~Van~Gool, ``Untrimmednets for weakly supervised action recognition and detection,'' in \emph{Proceedings of the IEEE conference on Computer Vision and Pattern Recognition}, 2017, pp. 4325--4334.

\bibitem{a2cl-pt}
K.~Min and J.~J. Corso, ``Adversarial background-aware loss for weakly-supervised temporal activity localization,'' in \emph{Computer Vision - {ECCV} 2020 - 16th European Conference, Glasgow, UK, August 23-28, 2020, Proceedings, Part {XIV}}, vol. 12359, 2020, pp. 283--299.

\bibitem{narayan20193c}
S.~Narayan, H.~Cholakkal, F.~S. Khan, and L.~Shao, ``3c-net: Category count and center loss for weakly-supervised action localization,'' in \emph{Proceedings of the IEEE/CVF International Conference on Computer Vision}, 2019, pp. 8679--8687.

\bibitem{paulWTALCWeaklySupervisedTemporal2018}
S.~Paul, S.~Roy, and A.~K. {Roy-Chowdhury}, ``W-{{TALC}}: {{Weakly-Supervised Temporal Activity Localization}} and {{Classification}},'' in \emph{Computer {{Vision}} \textendash{} {{ECCV}} 2018}, V.~Ferrari, M.~Hebert, C.~Sminchisescu, and Y.~Weiss, Eds., vol. 11208.\hskip 1em plus 0.5em minus 0.4em\relax {Cham}: {Springer International Publishing}, 2018, pp. 588--607.

\bibitem{sunSlowMotionMatters2022}
W.~Sun, R.~Su, Q.~Yu, and D.~Xu, ``Slow motion matters: A slow motion enhanced network for weakly supervised temporal action localization,'' \emph{IEEE Transactions on Circuits and Systems for Video Technology}, vol.~33, no.~1, pp. 354--366, 2022.

\bibitem{wang2021exploring}
B.~Wang, X.~Zhang, and Y.~Zhao, ``Exploring sub-action granularity for weakly supervised temporal action localization,'' \emph{IEEE Transactions on Circuits and Systems for Video Technology}, vol.~32, no.~4, pp. 2186--2198, 2021.

\bibitem{islam2021hybrid}
A.~Islam, C.~Long, and R.~Radke, ``A hybrid attention mechanism for weakly-supervised temporal action localization,'' in \emph{Proceedings of the AAAI Conference on Artificial Intelligence}, vol.~35, no.~2, 2021, pp. 1637--1645.

\bibitem{ju2022adaptive}
C.~Ju, P.~Zhao, S.~Chen, Y.~Zhang, X.~Zhang, Y.~Wang, and Q.~Tian, ``Adaptive mutual supervision for weakly-supervised temporal action localization,'' \emph{IEEE Transactions on Multimedia}, 2022.

\bibitem{zeng2019breaking}
R.~Zeng, C.~Gan, P.~Chen, W.~Huang, Q.~Wu, and M.~Tan, ``Breaking winner-takes-all: Iterative-winners-out networks for weakly supervised temporal action localization,'' \emph{IEEE Transactions on Image Processing}, vol.~28, no.~12, pp. 5797--5808, 2019.

\bibitem{yang2021multi}
W.~Yang, T.~Zhang, Z.~Mao, Y.~Zhang, Q.~Tian, and F.~Wu, ``Multi-scale structure-aware network for weakly supervised temporal action detection,'' \emph{IEEE Transactions on Image Processing}, vol.~30, pp. 5848--5861, 2021.

\bibitem{song2022slow}
P.~Song and C.~Zhao, ``Slow down to go better: A survey on slow feature analysis,'' \emph{IEEE Transactions on Neural Networks and Learning Systems}, 2022.

\bibitem{kumar2017hide}
K.~Kumar~Singh and Y.~Jae~Lee, ``Hide-and-seek: Forcing a network to be meticulous for weakly-supervised object and action localization,'' in \emph{Proceedings of the IEEE International Conference on Computer Vision}, 2017, pp. 3524--3533.

\bibitem{min2020adversarial}
K.~Min and J.~J. Corso, ``Adversarial background-aware loss for weakly-supervised temporal activity localization,'' in \emph{Computer Vision--ECCV 2020: 16th European Conference, Glasgow, UK, August 23--28, 2020, Proceedings, Part XIV 16}.\hskip 1em plus 0.5em minus 0.4em\relax Springer, 2020, pp. 283--299.

\bibitem{pardoRefineLocIterativeRefinement2021}
A.~Pardo, H.~Alwassel, F.~C. Heilbron, A.~Thabet, and B.~Ghanem, ``{{RefineLoc}}: {{Iterative Refinement}} for {{Weakly-Supervised Action Localization}},'' in \emph{2021 {{IEEE Winter Conference}} on {{Applications}} of {{Computer Vision}} ({{WACV}})}.\hskip 1em plus 0.5em minus 0.4em\relax {Waikoloa, HI, USA}: {IEEE}, Jan. 2021, pp. 3318--3327.

\bibitem{chen2021exploring}
X.~Chen and K.~He, ``Exploring simple siamese representation learning,'' in \emph{Proceedings of the IEEE/CVF conference on computer vision and pattern recognition}, 2021, pp. 15\,750--15\,758.

\bibitem{caron2018deep}
M.~Caron, P.~Bojanowski, A.~Joulin, and M.~Douze, ``Deep clustering for unsupervised learning of visual features,'' in \emph{Proceedings of the European conference on computer vision (ECCV)}, 2018, pp. 132--149.

\bibitem{chen2020simple}
T.~Chen, S.~Kornblith, M.~Norouzi, and G.~Hinton, ``A simple framework for contrastive learning of visual representations,'' in \emph{International conference on machine learning}.\hskip 1em plus 0.5em minus 0.4em\relax PMLR, 2020, pp. 1597--1607.

\bibitem{he2020momentum}
K.~He, H.~Fan, Y.~Wu, S.~Xie, and R.~Girshick, ``Momentum contrast for unsupervised visual representation learning,'' in \emph{Proceedings of the IEEE/CVF conference on computer vision and pattern recognition}, 2020, pp. 9729--9738.

\bibitem{wu2018unsupervised}
Z.~Wu, Y.~Xiong, S.~X. Yu, and D.~Lin, ``Unsupervised feature learning via non-parametric instance discrimination,'' in \emph{Proceedings of the IEEE conference on computer vision and pattern recognition}, 2018, pp. 3733--3742.

\bibitem{liu2022multi}
Z.~Liu, C.~Zhao, Y.~Lu, Y.~Jiang, and J.~Yan, ``Multi-scale graph learning for ovarian tumor segmentation from ct images,'' \emph{Neurocomputing}, vol. 512, pp. 398--407, 2022.

\bibitem{khosla2020supervised}
P.~Khosla, P.~Teterwak, C.~Wang, A.~Sarna, Y.~Tian, P.~Isola, A.~Maschinot, C.~Liu, and D.~Krishnan, ``Supervised contrastive learning,'' \emph{Advances in neural information processing systems}, vol.~33, pp. 18\,661--18\,673, 2020.

\bibitem{liu2022semi}
Z.~Liu and C.~Zhao, ``Semi-supervised medical image segmentation via geometry-aware consistency training,'' \emph{arXiv preprint arXiv:2202.06104}, 2022.

\bibitem{tao2022improved}
L.~Tao, X.~Wang, and T.~Yamasaki, ``An improved inter-intra contrastive learning framework on self-supervised video representation,'' \emph{IEEE Transactions on Circuits and Systems for Video Technology}, vol.~32, no.~8, pp. 5266--5280, 2022.

\bibitem{chen2022consistent}
Z.~Chen, K.-Y. Lin, and W.-S. Zheng, ``Consistent intra-video contrastive learning with asynchronous long-term memory bank,'' \emph{IEEE Transactions on Circuits and Systems for Video Technology}, vol.~33, no.~3, pp. 1168--1180, 2022.

\bibitem{huang2021self}
J.~Huang, Y.~Huang, Q.~Wang, W.~Yang, and H.~Meng, ``Self-supervised representation learning for videos by segmenting via sampling rate order prediction,'' \emph{IEEE Transactions on Circuits and Systems for Video Technology}, vol.~32, no.~6, pp. 3475--3489, 2021.

\bibitem{xu2022x}
B.~Xu, X.~Shu, and Y.~Song, ``X-invariant contrastive augmentation and representation learning for semi-supervised skeleton-based action recognition,'' \emph{IEEE Transactions on Image Processing}, vol.~31, pp. 3852--3867, 2022.

\bibitem{shu2022multi}
X.~Shu, B.~Xu, L.~Zhang, and J.~Tang, ``Multi-granularity anchor-contrastive representation learning for semi-supervised skeleton-based action recognition,'' \emph{IEEE Transactions on Pattern Analysis and Machine Intelligence}, 2022.

\bibitem{xu2023pyramid}
B.~Xu and X.~Shu, ``Pyramid self-attention polymerization learning for semi-supervised skeleton-based action recognition,'' \emph{arXiv preprint arXiv:2302.02327}, 2023.

\bibitem{xu2023spatiotemporal}
B.~Xu, X.~Shu, J.~Zhang, G.~Dai, and Y.~Song, ``Spatiotemporal decouple-and-squeeze contrastive learning for semisupervised skeleton-based action recognition,'' \emph{IEEE Transactions on Neural Networks and Learning Systems}, 2023.

\bibitem{luoActionUnitMemory2021}
W.~Luo, T.~Zhang, W.~Yang, J.~Liu, T.~Mei, F.~Wu, and Y.~Zhang, ``Action {{Unit Memory Network}} for {{Weakly Supervised Temporal Action Localization}},'' in \emph{2021 {{IEEE}}/{{CVF Conference}} on {{Computer Vision}} and {{Pattern Recognition}} ({{CVPR}})}.\hskip 1em plus 0.5em minus 0.4em\relax {Nashville, TN, USA}: {IEEE}, Jun. 2021, pp. 9964--9974.

\bibitem{I3D}
J.~Carreira and A.~Zisserman, ``Quo vadis, action recognition? a new model and the kinetics dataset,'' in \emph{proceedings of the IEEE Conference on Computer Vision and Pattern Recognition}, 2017, pp. 6299--6308.

\bibitem{detr}
N.~Carion, F.~Massa, G.~Synnaeve, N.~Usunier, A.~Kirillov, and S.~Zagoruyko, ``End-to-end object detection with transformers,'' in \emph{Computer Vision - {ECCV} 2020 - 16th European Conference, Glasgow, UK, August 23-28, 2020, Proceedings, Part {I}}, vol. 12346, 2020, pp. 213--229.

\bibitem{MIL}
T.~G. Dietterich, R.~H. Lathrop, and T.~Lozano-P{\'e}rez, ``Solving the multiple instance problem with axis-parallel rectangles,'' \emph{Artificial intelligence}, vol.~89, no. 1-2, pp. 31--71, 1997.

\bibitem{zhang2018generalized}
Z.~Zhang and M.~Sabuncu, ``Generalized cross entropy loss for training deep neural networks with noisy labels,'' \emph{Advances in neural information processing systems}, vol.~31, 2018.

\bibitem{ghosh2015making}
A.~Ghosh, N.~Manwani, and P.~Sastry, ``Making risk minimization tolerant to label noise,'' \emph{Neurocomputing}, vol. 160, pp. 93--107, 2015.

\bibitem{chollet2017xception}
F.~Chollet, ``Xception: Deep learning with depthwise separable convolutions,'' in \emph{Proceedings of the IEEE conference on computer vision and pattern recognition}, 2017, pp. 1251--1258.

\bibitem{transformer}
A.~Vaswani, N.~Shazeer, N.~Parmar, J.~Uszkoreit, L.~Jones, A.~N. Gomez, L.~Kaiser, and I.~Polosukhin, ``Attention is all you need,'' in \emph{Advances in Neural Information Processing Systems 30: Annual Conference on Neural Information Processing Systems 2017, December 4-9, 2017, Long Beach, CA, {USA}}, 2017, pp. 5998--6008.

\bibitem{xu2020privileged}
C.~Xu, Q.~Li, J.~Ge, J.~Gao, X.~Yang, C.~Pei, F.~Sun, J.~Wu, H.~Sun, and W.~Ou, ``Privileged features distillation at taobao recommendations,'' in \emph{Proceedings of the 26th ACM SIGKDD International Conference on Knowledge Discovery \& Data Mining}, 2020, pp. 2590--2598.

\bibitem{vapnik2015privilegedlearning}
V.~Vapnik, R.~Izmailov \emph{et~al.}, ``Learning using privileged information: similarity control and knowledge transfer.'' \emph{J. Mach. Learn. Res.}, vol.~16, no.~1, pp. 2023--2049, 2015.

\bibitem{kendall2017uncertainties}
A.~Kendall and Y.~Gal, ``What uncertainties do we need in bayesian deep learning for computer vision?'' \emph{Advances in neural information processing systems}, vol.~30, 2017.

\bibitem{gawlikowski2023survey}
J.~Gawlikowski, C.~R.~N. Tassi, M.~Ali, J.~Lee, M.~Humt, J.~Feng, A.~Kruspe, R.~Triebel, P.~Jung, R.~Roscher \emph{et~al.}, ``A survey of uncertainty in deep neural networks,'' \emph{Artificial Intelligence Review}, pp. 1--77, 2023.

\bibitem{smith2018understanding}
L.~Smith and Y.~Gal, ``Understanding measures of uncertainty for adversarial example detection,'' \emph{arXiv preprint arXiv:1803.08533}, 2018.

\bibitem{lin2019bmn}
T.~Lin, X.~Liu, X.~Li, E.~Ding, and S.~Wen, ``Bmn: Boundary-matching network for temporal action proposal generation,'' in \emph{Proceedings of the IEEE/CVF international conference on computer vision}, 2019, pp. 3889--3898.

\bibitem{xu2020gtad}
M.~Xu, C.~Zhao, D.~S. Rojas, A.~Thabet, and B.~Ghanem, ``G-tad: Sub-graph localization for temporal action detection,'' in \emph{Proceedings of the IEEE/CVF conference on computer vision and pattern recognition}, 2020, pp. 10\,156--10\,165.

\bibitem{TadTR}
X.~Liu, Q.~Wang, Y.~Hu, X.~Tang, S.~Zhang, S.~Bai, and X.~Bai, ``End-to-end temporal action detection with transformer,'' \emph{{IEEE} Trans. Image Process.}, vol.~31, pp. 5427--5441, 2022.

\bibitem{STAR}
Y.~Xu, C.~Zhang, Z.~Cheng, J.~Xie, Y.~Niu, S.~Pu, and F.~Wu, ``Segregated temporal assembly recurrent networks for weakly supervised multiple action detection,'' in \emph{Proceedings of the AAAI conference on artificial intelligence}, vol.~33, no.~01, 2019, pp. 9070--9078.

\bibitem{shou2018autoloc}
Z.~Shou, H.~Gao, L.~Zhang, K.~Miyazawa, and S.-F. Chang, ``Autoloc: Weakly-supervised temporal action localization in untrimmed videos,'' in \emph{Proceedings of the European Conference on Computer Vision (ECCV)}, 2018, pp. 154--171.

\bibitem{nguyenWeaklySupervisedAction2018}
P.~Nguyen, B.~Han, T.~Liu, and G.~Prasad, ``Weakly {{Supervised Action Localization}} by {{Sparse Temporal Pooling Network}},'' in \emph{2018 {{IEEE}}/{{CVF Conference}} on {{Computer Vision}} and {{Pattern Recognition}}}.\hskip 1em plus 0.5em minus 0.4em\relax {Salt Lake City, UT, USA}: {IEEE}, Jun. 2018, pp. 6752--6761.

\bibitem{dgam}
B.~Shi, Q.~Dai, Y.~Mu, and J.~Wang, ``Weakly-supervised action localization by generative attention modeling,'' in \emph{Proceedings of the IEEE/CVF Conference on Computer Vision and Pattern Recognition}, 2020, pp. 1009--1019.

\bibitem{TSCN}
Y.~Zhai, L.~Wang, W.~Tang, Q.~Zhang, J.~Yuan, and G.~Hua, ``Two-stream consensus network for weakly-supervised temporal action localization,'' in \emph{Computer Vision--ECCV 2020: 16th European Conference, Glasgow, UK, August 23--28, 2020, Proceedings, Part VI 16}.\hskip 1em plus 0.5em minus 0.4em\relax Springer, 2020, pp. 37--54.

\bibitem{huangModelingSubActionsWeakly2021}
L.~Huang, Y.~Huang, W.~Ouyang, and L.~Wang, ``Modeling {{Sub-Actions}} for {{Weakly Supervised Temporal Action Localization}},'' \emph{IEEE Transactions on Image Processing}, vol.~30, pp. 5154--5167, 2021.

\bibitem{narayan2021d2}
S.~Narayan, H.~Cholakkal, M.~Hayat, F.~S. Khan, M.-H. Yang, and L.~Shao, ``D2-net: Weakly-supervised action localization via discriminative embeddings and denoised activations,'' in \emph{Proceedings of the IEEE/CVF International Conference on Computer Vision}, 2021, pp. 13\,608--13\,617.

\bibitem{hongCrossmodalConsensusNetwork2021}
F.-T. Hong, J.-C. Feng, D.~Xu, Y.~Shan, and W.-S. Zheng, ``Cross-modal {{Consensus Network}} for {{Weakly Supervised Temporal Action Localization}},'' Jul. 2021.

\bibitem{gao2022fine}
J.~Gao, M.~Chen, and C.~Xu, ``Fine-grained temporal contrastive learning for weakly-supervised temporal action localization,'' in \emph{Proceedings of the IEEE/CVF Conference on Computer Vision and Pattern Recognition}, 2022, pp. 19\,999--20\,009.

\bibitem{liExploringDenoisedCrossvideo2022}
J.~Li, T.~Yang, W.~Ji, J.~Wang, and L.~Cheng, ``Exploring {{Denoised Cross-video Contrast}} for {{Weakly-supervised Temporal Action Localization}},'' in \emph{2022 {{IEEE}}/{{CVF Conference}} on {{Computer Vision}} and {{Pattern Recognition}} ({{CVPR}})}.\hskip 1em plus 0.5em minus 0.4em\relax {New Orleans, LA, USA}: {IEEE}, Jun. 2022, pp. 19\,882--19\,892.

\bibitem{ASL}
J.~Ma, S.~K. Gorti, M.~Volkovs, and G.~Yu, ``Weakly supervised action selection learning in video,'' in \emph{Proceedings of the IEEE/CVF Conference on Computer Vision and Pattern Recognition}, 2021, pp. 7587--7596.

\bibitem{HAAN}
Z.~Li, L.~He, and H.~Xu, ``Weakly-supervised temporal action detection for fine-grained videos with hierarchical atomic actions,'' in \emph{Computer Vision - {ECCV} 2022 - 17th European Conference, Tel Aviv, Israel, October 23-27, 2022, Proceedings, Part {X}}, vol. 13670, 2022, pp. 567--584.

\bibitem{attention}
A.~Vaswani, N.~Shazeer, N.~Parmar, J.~Uszkoreit, L.~Jones, A.~N. Gomez, L.~Kaiser, and I.~Polosukhin, ``Attention is all you need,'' in \emph{Advances in Neural Information Processing Systems 30: Annual Conference on Neural Information Processing Systems 2017, December 4-9, 2017, Long Beach, CA, {USA}}, 2017, pp. 5998--6008.

\bibitem{adam}
D.~P. Kingma and J.~Ba, ``Adam: {A} method for stochastic optimization,'' in \emph{3rd International Conference on Learning Representations, {ICLR} 2015, San Diego, CA, USA, May 7-9, 2015, Conference Track Proceedings}, Y.~Bengio and Y.~LeCun, Eds., 2015.

\bibitem{DBSCAN}
M.~Ester, H.-P. Kriegel, J.~Sander, X.~Xu \emph{et~al.}, ``A density-based algorithm for discovering clusters in large spatial databases with noise.'' in \emph{kdd}, vol.~96, no.~34, 1996, pp. 226--231.

\bibitem{asano2019self}
Y.~M. Asano, C.~Rupprecht, and A.~Vedaldi, ``Self-labelling via simultaneous clustering and representation learning,'' in \emph{8th International Conference on Learning Representations, {ICLR} 2020, Addis Ababa, Ethiopia, April 26-30, 2020}, 2020.

\bibitem{li2019expectation}
X.~Li, Z.~Zhong, J.~Wu, Y.~Yang, Z.~Lin, and H.~Liu, ``Expectation-maximization attention networks for semantic segmentation,'' in \emph{Proceedings of the IEEE/CVF International Conference on Computer Vision}, 2019, pp. 9167--9176.

\end{thebibliography}

\begin{IEEEbiography}[{\includegraphics[width=1in,height=1.25in,clip,keepaspectratio]{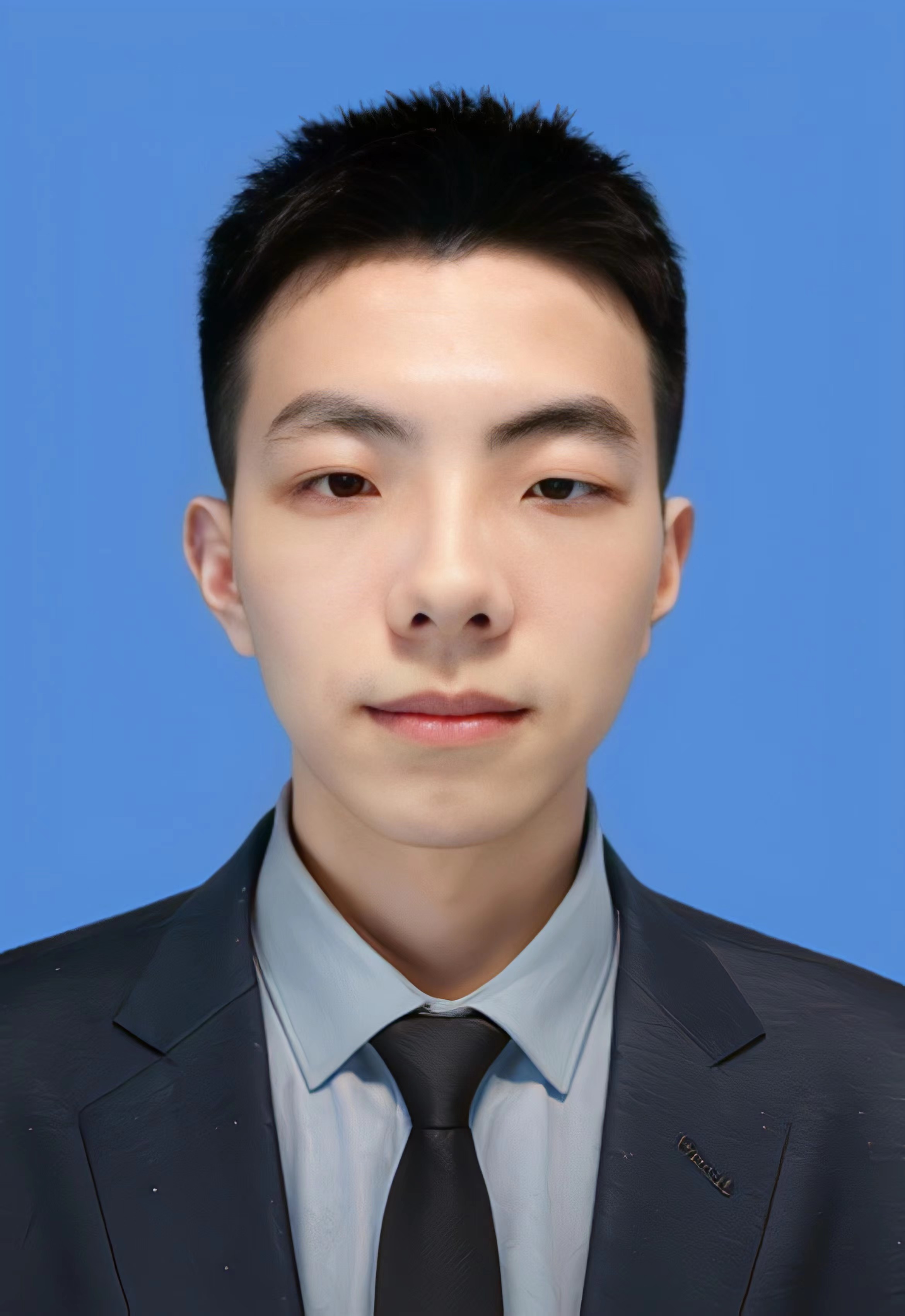}}]{Songchun Zhang} received the B.S. degree in electrical engineering and automation from Hunan University, Changsha, China, in 2022. He is currently pursuing the M.S. degree with the College of Control Science and Engineering, Zhejiang University, Hangzhou, China. He has wide research interests mainly including computer vision, video understanding, and 3D generative models.
\end{IEEEbiography}

\begin{IEEEbiography}[{\includegraphics[width=1in,height=1.25in,clip,keepaspectratio]{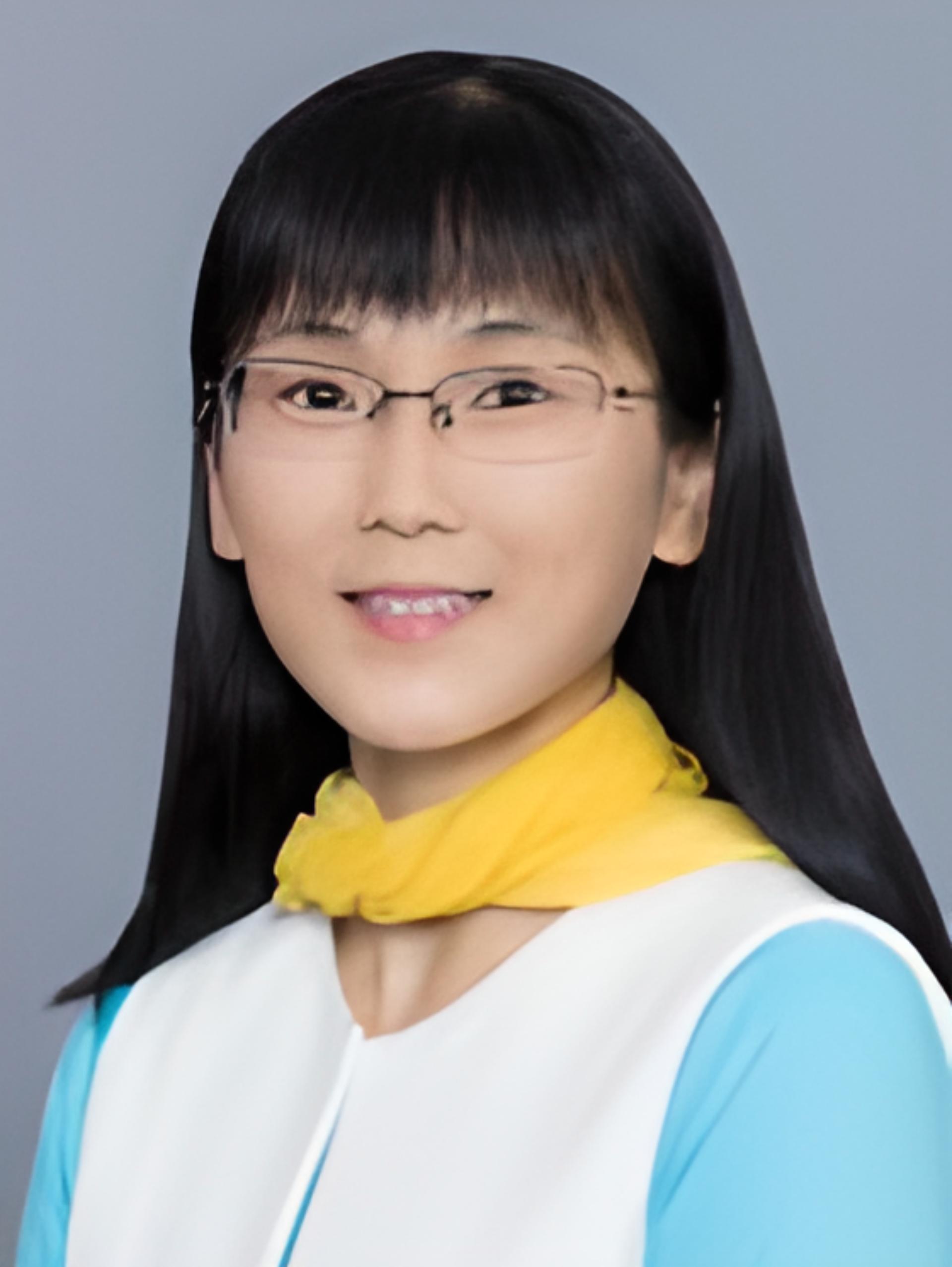}}]{Chunhui Zhao}  (Senior Member, IEEE) received the Ph.D. degree from Northeastern University, Shenyang, China, in 2009. 

From 2009 to 2012, she was a Post-Doctoral Fellow with The Hong Kong University of Science and Technology, Hong Kong, and the University of California at Santa Barbara, Santa Barbara, CA, USA. Since January 2012, she has been a Professor with the College of Control Science and Engineering, Zhejiang University, Hangzhou, China. She has authored or coauthored more than 170 papers in peer-reviewed international journals. Her research interests include statistical machine learning and data mining for industrial applications. Dr. Zhao has served as a Senior Editor for Journal of Process Control and an Associate Editor for two international journals, including Control Engineering Practice and Neurocomputing.
\end{IEEEbiography}

\end{document}